\documentclass{article}

\usepackage{microtype}
\usepackage{graphicx}
\usepackage{subcaption}
\usepackage{multirow}
\usepackage{booktabs} 
\usepackage{soul}
\usepackage{tabularx}

\usepackage{hyperref}

\usepackage[preprint]{icml2023}

\usepackage{amsmath}
\usepackage{amssymb}
\usepackage{mathtools}
\usepackage{amsthm}
\usepackage{amssymb}
\usepackage{amsmath,amsfonts}
\usepackage{amsopn,bm,mathtools}

\usepackage{nicefrac}       % compact symbols for 1/2, etc.
\newcommand{\ProbOpr}[1]{\mathbb{#1}}
\newcommand{\expect}[2]{%
\ifthenelse{\equal{#2}{}}{\ProbOpr{E}_{#1}}
{\ifthenelse{\equal{#1}{}}{\ProbOpr{E}\left[#2\right]}{\ProbOpr{E}_{#1}\left[#2\right]}}} % Expectation: syntax: E{1}{2} = E_1[2], E{}{2}=E[2], E{1}{} = E_1
\newcommand{\var}[2]{%
\ifthenelse{\equal{#2}{}}{\ProbOpr{VAR}_{#1}}
{\ifthenelse{\equal{#1}{}}{\ProbOpr{VAR}\left[#2\right]}{\ProbOpr{VAR}_{#1}\left[#2\right]}}} % Expectation: syntax: V{1}{2} = V_1[2], V{}{2}=V[2], V{1}{} = V_1
\usepackage{xspace}
\makeatletter
\DeclareRobustCommand\onedot{\futurelet\@let@token\@onedot}
\def\@onedot{\ifx\@let@token.\else.\null\fi\xspace}
\def\eg{\emph{e.g}\onedot} 
\def\ie{\emph{i.e}\onedot}

\def\etal{\emph{et al}\onedot}
\makeatother

\usepackage{mathtools}

\usepackage{listings}

\graphicspath{{../figures/}{../}}

\newcommand{\eat}[1]{}

\usepackage{makecell}

\usepackage{enumitem}
\usepackage{pbox}
\usepackage[font={small}]{caption}

\usepackage[capitalize,noabbrev]{cleveref}

\theoremstyle{plain}

\theoremstyle{definition}

\theoremstyle{remark}

\usepackage[textsize=tiny]{todonotes}

\def\STAIRs{STAIR\textsubscript{\textsc{single-stage}}\xspace}
\def\STAIRm{STAIR }

\icmltitlerunning{STAIR: Learning Sparse Text and Image Representation in Grounded Tokens}

\begin{document}

\twocolumn[
\icmltitle{STAIR: Learning Sparse Text and Image Representation in Grounded Tokens}

\icmlsetsymbol{equal}{*}

\begin{icmlauthorlist}
\icmlauthor{Chen Chen}{comp}
\icmlauthor{Bowen Zhang}{comp}
\icmlauthor{Liangliang Cao}{comp}
\icmlauthor{Jiguang Shen}{comp}
\icmlauthor{Tom Gunter}{comp}
\icmlauthor{Albin Madappally Jose}{comp}
\icmlauthor{Alexander Toshev}{comp}
\icmlauthor{Jonathon Shlens}{comp}
\icmlauthor{Ruoming Pang}{comp}
\icmlauthor{Yinfei Yang}{comp}
\end{icmlauthorlist}

\icmlaffiliation{comp}{Apple Inc, Cupertino, US}

\icmlcorrespondingauthor{Chen Chen}{chen\_chen999@apple.com}
\icmlcorrespondingauthor{Bowen Zhang}{bowen\_zhang4@apple.com}
\icmlcorrespondingauthor{Yinfei Yang}{yinfeiy@apple.com}

\icmlkeywords{Machine Learning, ICML}

\vskip 0.3in
]

\printAffiliationsAndNotice{} 

%auto-ignore
\begin{abstract}
    Image and text retrieval is one of the foundational tasks in the vision and language domain with multiple real-world applications. State-of-the-art approaches, \eg CLIP~\cite{clip}, ALIGN~\cite{jia2021scaling}, represent images and texts as dense embeddings and calculate the similarity in the dense embedding space as the matching score. On the other hand, sparse semantic features like bag-of-words models are more interpretable, but believed to suffer from inferior accuracy than dense representations. In this work, we show that it is possible to build a sparse semantic representation that is as powerful as, or even better than, dense presentations. We extend the CLIP model and build a sparse text and image representation (STAIR), where the image and text are mapped to a sparse token space. Each token in the space is a (sub-)word in the vocabulary, which is not only interpretable but also easy to integrate with existing information retrieval systems. STAIR model significantly outperforms a CLIP model with +$4.9\%$ and +$4.3\%$ absolute Recall@1 improvement on COCO-5k text$\rightarrow$image and image$\rightarrow$text retrieval respectively. It also achieved better performance on both of ImageNet zero-shot and linear probing compared to CLIP.

\end{abstract}

\section{Introduction}
\label{sec:intro}

%auto-ignore
Learning high-quality and performant representations from large-scale image-text data has been the subject of intense study in the past few years. Many applications benefit from an improved vision-language representation, including image-text retrieval~\cite{coco_captions,plummer2015flickr30k}, VQA~\cite{vqa,johnson2017clevr}, and image captioning~\cite{vinyals2015_captioning}.
Recent advances have been fueled by a renewed interest in contrastive learning.
State-of-the-art contrastive learning models such as CLIP~\cite{clip} and ALIGN~\cite{jia2021scaling} use image and text specific encoders to embed images and text in a joint embedding space.
The model is trained to improve the cosine similarity for the aligned image-text pairs and dissimilarity for the unmatched ones. Such contrastive learning models can achieve state-of-the-art fine-tuning performance as well as strong zero-shot generalization results on image-text retrieval, VQA, and image classification.

\begin{figure}[t]
\begin{center}
\includegraphics[width=0.41\textwidth]{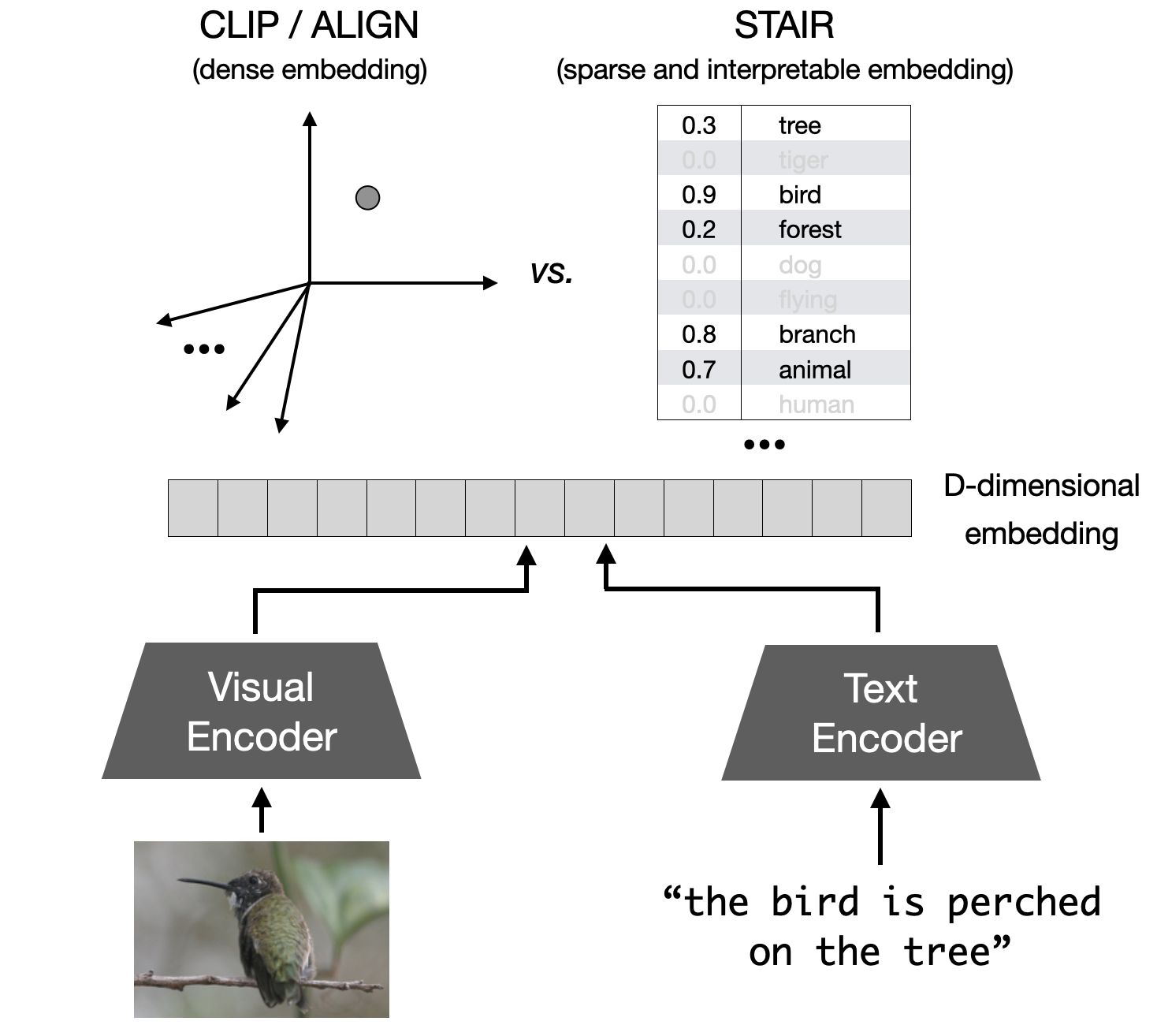}
\end{center}
\vspace{-0.2cm}
\caption{{\bf Learning a sparse and interpretable embedding for a vision-language model.} CLIP \cite{clip} and ALIGN \cite{jia2021scaling} learn a dense $D$-dimensional embedding. In contrast, the goal of STAIR is to learn a sparse and interpretable embedding in a high-dimensional space. Each dimension in the sparse embedding is a (sub-)word in a large dictionary in which the predicted non-negative scalar corresponds to the weight associated with the token.}
\label{fig:stair_diagram}
\vspace{-0.42cm}
\end{figure}

Despite the impressive benchmark performance, the dense embedding space is usually considered a black box which is hard and unintuitive to interpret the meaning of the embeddings.
Furthermore, as a model trained with retrieval~(contrastive) objective, deploying it to a real-world image-text retrieval product is challenging. Despite approximated nearest neighbor search~\cite{guo2020accelerating,faiss} can be used to retrieve from a dense embedding space, the cost can be high when scaling up to billions of images. 
Lastly, as the embeddings live in a dense space, traditional approaches developed by search engines, such as inverted index, cannot be deployed naively.
The difficulty of interoperability of the dense embeddings also makes it challenging to be combined with other retrieval features without additional training.

The computer vision field has made many efforts to explore sparse, interpretable representations of images. The bag-of-words and topic models \cite{csurka2004visual,sivic2005discovering,fei2005bayesian,lazebnik2006beyond} were widely used together with the SIFT descriptor \cite{lowe2004distinctive}, but later researchers found its performance is inferior to dense vectors \cite{lin2011large,sanchez2013image}. Another effort in the deep CNN era is to develop the deep visual-semantic embedding \cite{frome2013devise} using ImageNet topics, which is more interpretable but still fails to outperform dense representations \cite{faghri2017vse++}. 

We hypothesize that the gap between sparse semantic and dense embedding stems from at least two factors: (1) Previous works on semantic embedding do not effectively explore large-scale training to model the rich semantics in image-text space. (2) Most existing semantic embedding methods are built on a fixed vocabulary (e.g., several thousand concepts), which cannot handle out-of-vocabulary concepts. In this paper, we present a new model, named \textbf{STAIR}, and a multi-stage training recipe to learn a \textbf{S}parse \textbf{T}ext \textbf{A}nd \textbf{I}mage \textbf{R}epresentation to tackle the aforementioned challenges. We will show that our sparse representation can not only match but also outperform the state-of-the-art of image-text representations. 

Inspired by the recent success of the sparse representation in information retrieval  field~\cite{bai2020sparterm,formal2021splade}, 
STAIR encodes the image and text into a sparse and grounded token space, as illustrated in Figure~\ref{fig:stair_diagram}. Concretely, images and text are mapped to sparse embeddings that are grounded to actual (sub-)words with non-negative weights. The sparse embedding is straightforward to \textit{1) interpret and understand the model's prediction; 2) conduct a large-scale retrieval using an efficient inverted index approach; 3) combine with other text features by simply biasing the token weights in the sparse token space.} The proposed multi-stage training is critical to the grounding of the model.

We compare the STAIR model and a CLIP model trained using the same model architecture and dataset. Experiment results show the STAIR model significantly outperforms the CLIP model on image-text retrieval tasks, with +4.9\% and +4.3\% recall@1 on COCO-5K text$\rightarrow$image and image$\rightarrow$text retrieval respectively. STAIR models can also achieve similar or better performance on zero-shot classification and linear probing tasks including ImageNet. 

Furthermore, our sparse embedding is easier for humans to interpret. We define an interpretable space using BERT vocab to quantify the model interpretability. Experiments show \STAIRm is significantly more interpretable, with the Top-1 accuracy $32.9\%$~(STAIR) vs $13.7\%$~(CLIP) on ImageNet.

\section{Approach}
\label{sec:approach}

%auto-ignore
We begin with a review of a typical dual-encoder architecture to motivate our research.

\begin{figure*}[t]
\begin{center}
\includegraphics[width=0.9\textwidth]{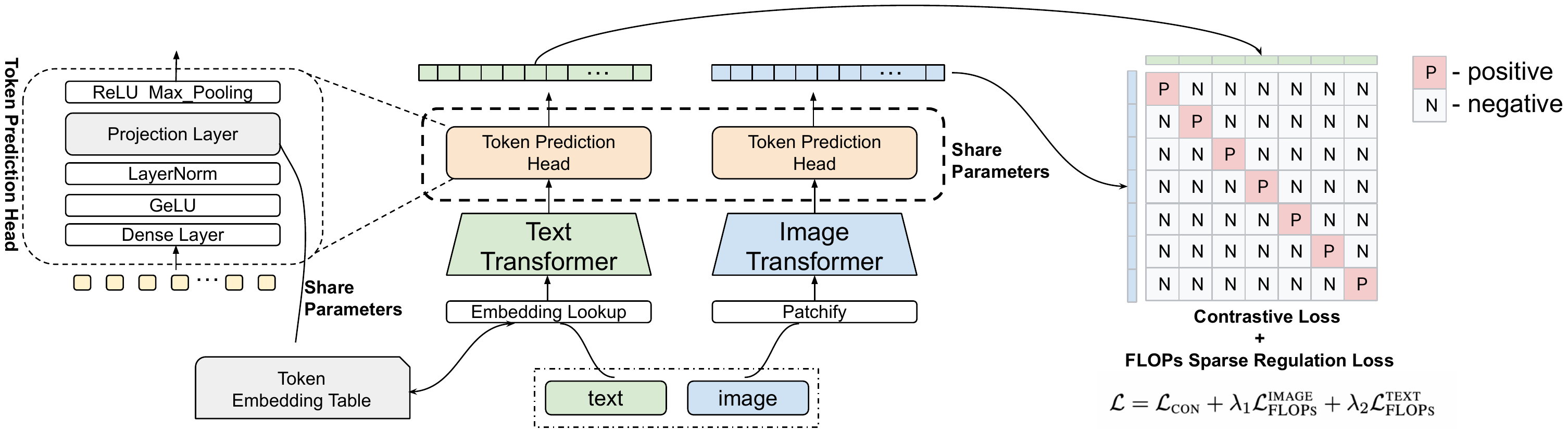}
\end{center}
\vspace{-0.2cm}
\caption{{\bf Diagram of \STAIRm architecture.} It employs a dual-encoder architecture. Different than dense models like CLIP or ALIGN, \STAIRm maps the dense embeddings to a sparse latent embedding space via a token prediction head. In addition to a regular contrastive loss to minimize an image-text matching loss, a FLOPs~\cite{flops} loss is added to encourage sparsity of image and text embeddings.}
\label{fig:stair}
\end{figure*}

\subsection{Dual-Encoder Contrastive Learning}
Given a dataset of image-text pairs $D=\{(x_i, y_i)\}$, where $x_i$ and $y_i$ represent the image and text respectively, a dual encoder learns a similarity $M(x, y)$, so that the aligned pair $(x_i, y_i)$ has higher similarity score than the unmatched pairs sampled from a negative sample set $D_i'$.

A dual-encoder architecture consists of an image encoder $E_{\textsc{image}}$ and a text encoder $E_{\textsc{text}}$. Each encoder $E$ has two components: 1) a standard, input-specific, neural network $f(\cdot)$ and 2) a projection head $g(\cdot)$ that maps the features as embeddings in a joint dense embedding space:
\begin{equation}
 E(\cdot)=g(f(\cdot))
  \label{eq:encoder}
\end{equation}
State-of-the-art approaches usually set $f(\cdot)$ as a transformer model and $g(\cdot)$ as a pooling layer.

Given an image $x$ and text $y$ inputs, their similarity is measured by the cosine similarity:
\begin{equation}
    M(x,y) = \frac{E(x)^T E(y)}{\lVert E(x) \rVert \lVert E(y) \rVert}
\end{equation}

It is used to train the dual encoder by minimizing the contrastive loss $\mathcal{L}_{\textsc{con}}$: 

\begin{equation}
\label{eq:con}
    \mathcal{L}_{\textsc{con}} =  - \frac{1}{|D|}\sum_i \log \frac{e^{M(x_i,y_i)/T}}{\sum_{(x', y')\in D_i}e^{M(x', y')/T}}
\end{equation}
where $D_i = D_i' \cup \{(x_i, y_i)\}$ denotes the positive pair and its negative set for each pair, and $T$ is the softmax temperature.

\subsection{STAIR}
\label{sssec:stair}

Following CLIP~\cite{clip}, \STAIRm employs a dual-encoder architecture. As illustrated in Figure \ref{fig:stair}, it contains a pair of image and text encoders with sequential feature extraction network $f(\cdot)$ and projection head $g(\cdot)$. In particular, the dense project head $g(\cdot)$ is replaced with a \textbf{Token Projection Head}, which maps the representation to a sparse embedding space. A vocabulary $V$ is used as the basis of embedding space for the purpose of interpretability.

The token projection head $g(\cdot)$ consists of two components: 1) a mapping function that maps the input sequence $h$ to a sequence of weights for each token $j$ in the vocabulary space $V$ and 2) a pooling layer that summarizes the sequence as sparse embedding in the vocabulary space $V$. We reuse the BERT~\cite{kenton2019bert} masked language model (MLM) prediction head $p(\cdot)$ as the mapping function:

\begin{equation}
\label{eq:gmap}
    p(h_j) = e \cdot \textsc{transform}(h_j) + b
\end{equation}
where $h_j = f(\cdot)$ corresponds to the $j^{\text{th}}$ token in the sequence of the feature extraction network output, the $\textsc{transform}$ function consists of a FC layer with \textsc{GeLU} activation and a \textsc{Layer Norm} layer, and $e$ and $b$ are the linear mapping and bias in MLM prediction head that maps the transformed embedding to vocabulary space. We tie the weights of linear mapping layer $e$ with the text tower token embedding table. We believe that this step is important to associate the sparse embedding to actual text tokens in $V$.

Following \cite{formal2021splade,formal2021spladev2}, we aggregate the weight of token $j$ to form the sparse embedding $\textsc{enc}$:
\begin{equation}
    \textsc{enc} = \log(1+\textsc{ReLU}(\underset{j}{\max}(p(h_j))))
     \label{eq:log_max_pooling}
\end{equation}
The \textsc{ReLU} activation ensures that each token weight is non-negative and adding the log function empirically achieves better performance by suppressing overwhelmingly large weights~\cite{sparta}. 
After the token projection head, the image embedding $\textsc{enc}^{\textsc{image}}$ and text embedding $\textsc{enc}^{\textsc{text}}$ are sparse vectors living in a $|V|$-dimensional space defined by the vocabulary.

To further encourage sparsity, we follow ~\cite{flops} to introduce the FLOPs regularization loss such that only a small number of token embeddings in $V$ are non-zeros:

\begin{equation}
    \mathcal{L}_{\textsc{FLOPs}} = \sum_{k \in V}(\frac{1}{N}\sum_{i=1}^N\textsc{enc}_{k}^{(i)})^2 
\end{equation}

By combining the contrastive loss and the FLOPs regularization loss, the \STAIRm model is optimized by:
\begin{equation}
    \mathcal{L} = \mathcal{L}_{\textsc{con}} + \lambda_1 \mathcal{L}_{\textsc{FLOPs}}^{\textsc{image}} + \lambda_2  \mathcal{L}_{\textsc{FLOPs}}^{\textsc{text}}
    \label{eq:final_training_objectives}
\end{equation}
where $\lambda_1$ and $\lambda_2$ are FLOPs regularization weights for image and text embeddings correspondingly.

\section{Training Details}
\label{sec:training}

%auto-ignore
\begin{figure}[t]
\begin{center}
\includegraphics[width=0.45\textwidth]{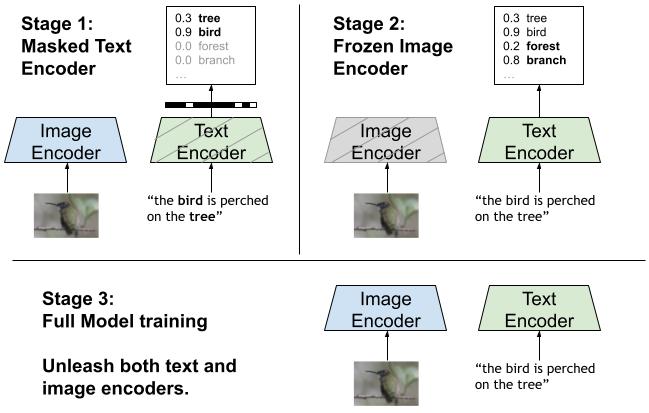}
\end{center}
\vspace{-0.5cm}
\caption{{\bf Training strategy for STAIR model.} (1) The text output is masked to only predict the weights of tokens that occur in the input. (2) The image encoder is frozen and the mask is removed from the text encoder. (3) All constraints are removed.}
\label{fig:stair_training}
\end{figure}

\begin{table*}[t]
\caption{{\bf Zero-shot text/image retrieval.} Reporting recall@K on Flickr30K and COCO.}
\vskip 0.15in
\small
  \centering
  \begin{tabular}{l|cccccc|cccccc}
    \toprule
    & \multicolumn{6}{c|}{\bf COCO 5K} & \multicolumn{6}{c}{\bf Flickr30K}\\
    & \multicolumn{3}{c}{text $\rightarrow$ image} & \multicolumn{3}{c|}{image $\rightarrow$ text} & \multicolumn{3}{c}{text $\rightarrow$ image} & \multicolumn{3}{c}{image $\rightarrow$ text}\\
    & R@1 & R@5 & R@10 & R@1  & R@5 & R@10 & R@1 & R@5 & R@10 & R@1 & R@5 & R@10\\
    \midrule

    \rule{-2pt}{8pt}
    CLIP & 36.2	& 62.2 & 72.2 &	53.4 & 78.3 & 85.6 & 63.0 &	86.7 & 92.5 & 79.6 & 95.5 &	98.1 \\
    \STAIRm & \bf{41.1} & \bf{65.4} & \bf{75.0} & \bf{57.7} & \bf{80.5} & \bf{87.3} & \bf{66.6} & \bf{88.7} & \bf{93.5} & \bf{81.2} & \bf{96.1} & \bf{98.4} \\
    \bottomrule
  \end{tabular}
  \label{tab:zero-shot retrieval}
\end{table*}

As noted in Section \ref{sssec:stair}, we expect the STAIR model to achieve two goals: 1) aligning text and images in the sparse embedding space; 2) grounding the sparse embedding dimension with human-understandable (sub-)word in the vocabulary. In other words, the image and text embeddings should eventually reside in a space spanned by basis vectors that ground to human-interpretable tokens.
However, in practice, we found that simply replacing the dense projection head with a token prediction head alone does not guarantee the $2^{\text{nd}}$ goal out of the box. This is because the images and text are two very different modalities with semantic gaps, while contrastive loss only encourages text/image alignment. As shown in Section \ref{sec:grounding}, the model learns a shortcut to reuse the less common tokens to bridge the gap across modalities. More details can be found in Appendix \ref{sec:single_stage}.

To address this issue, we propose a multi-stage training approach that sequentially bridges the discrepancy between the sparse embedding space and the interpretable space of vocabulary. The main steps are illustrated in Figure \ref{fig:stair_training}.

\paragraph{Stage 1: Training image embedding  with masked tokens} In the first stage, we co-train both the image and text encoders and apply a binary mask on the text embedding as illustrated in Figure \ref{fig:stair_training}. Formally, given the original text input \(y\), the masked text embedding is formulated as:
\begin{equation}
  \textsc{enc}^\textsc{text}_\textsc{mask}=g(f(y)) \cdot \textsc{mask}(y)
  \label{eq:bow}
\end{equation}
where $\textsc{mask}_i = 1$ if the $i^{\text{th}}$ token exists in the input sentence $y$ after tokenization, and $g(f(y))$ predicts the weights of the non-masked tokens. In this setting, the text embedding is forced to activate the tokens appearing from the original text input only, while ignoring others. By matching with the masked text embedding, the image encoder is learned to ground its image embedding on the tokens from the pairing text. Therefore, after the stage 1 training, the image embedding is living in the vocabulary's interpretable space.

\paragraph{Stage 2: Training with frozen image encoder} In this stage we focus on grounding the text embedding to the same interpretable space where the image embedding is trained to reside in from stage 1. The key idea is to let the image encoder teach the text encoder as a teacher model. As shown in Figure \ref{fig:stair_training}, we freeze the image encoder while training the text encoder to match the image embedding via contrastive loss. After stage 2 training, both image and text embeddings are in the same human-interpretable embedding space constructed by the vocabulary.

\paragraph{Stage 3: Fine-tuning both encoders} The first two-stage training provides a good foundation for both encoders to produce human-interpretable embeddings. However, the image and text encoders are tuned in a round-robin fashion, the parameters are not optimal for image-text matching. In stage 3, we boost the image-text matching performance by finetuning both encoders jointly.

\section{Experiments}
\label{sec:experiments}

%auto-ignore
\begin{table*}[t]
  \caption{{\bf Zero-shot classification accuracy.} Reporting the top-1 accuracy (\%) across 9 datasets.}
  \vskip 0.15in
\small
  \centering
  \begin{tabular}{l|ccccccccc}
    \toprule
    & ImageNet & Caltech-101 & CIFAR-100 & SVHN & DTD & OxPet & OxFlowers & Eurosat & RESISC45 \\
    \midrule
    CLIP & 65.1 &	82.3 &	63.2 &	42.0 &	53.6 &	85.8 &	67.7 &	\bf{52.4} &	\bf{64.3} \\
    \STAIRm & \bf{65.6} &	\bf{82.5} &	\bf{63.4} &	\bf{53.0} &	\bf{56.3} &	\bf{85.9} &	\bf{68.2} &	51.0 &	62.8 \\
    \bottomrule
  \end{tabular}

  \label{tab:zero-shot classification}
\end{table*}

\begin{table*}[t]
  \caption{{\bf Linear probe classification accuracy.} Reporting the top-1 accuracy (\%) across 9 datasets.}
\vskip 0.15in
\small
  \centering
  \begin{tabular}{l|ccccccccc}
    \toprule
    & ImageNet & Caltech-101 & CIFAR-100 & SVHN & DTD & OxPet & OxFlowers & Eurosat & RESISC45 \\
    \midrule
    CLIP & 77.0 &	94.9 &	76.7 &	65.8 &	80.6 &	88.8 &	97.6 &	95.1  &	92.6 \\
    % \STAIRs & 77.5 &	95.9 &	77.3 &	69.7 &	81.9 &	90.0 &	97.4 &	\bf{96.0} &	93.2  \\
    \STAIRm & \bf{78.1} &	\bf{96.1} &	\bf{77.7} &	\bf{72.5} &	\bf{82.7} &	\bf{91.7} &	\bf{98.2} &	\bf{95.9} &	\bf{93.9}  \\
    \bottomrule
  \end{tabular}

  %\caption{Accuracy@1(\%) of Zero-shot Classification Tasks}
  \label{tab:zero-shot classification}
\end{table*}

\subsection{Datasets}
\label{ssec:datasets}
Our dataset is a combination of internal and public datasets with 1.1B image-text pairs in total.
The public datasets consists of Conceptual Caption 3M (CC-3M)~\cite{sharma-etal-2018-conceptual} and Conceptual Captions 12M (CC-12M)~\cite{changpinyo2021cc12m}.
The internal image-text dataset consists of 1B image-text pairs, including a 134M clean licensed dataset (details in Appendix \ref{sec:ihqd}) along with a 971M noisy web-crawled dataset. The web-crawled dataset is mined using a similar approach described in ALIGN~\cite{jia2021scaling} and CLIP~\cite{clip}. We further filter the data by a public CLIP model \footnote{\url{https://huggingface.co/openai/clip-vit-base-patch16}}, where image-text pairs with similarity score $<$ 0.2 are removed.

\subsection{Configurations}
\label{ssec:config}
% We train three models for comparison, one dual-encoder baseline using dense representations (CLIP) and two \textsc{stair} models with same architecture where one takes the single-stage training strategy~(\STAIRs) while the other adopts the multi-stage training strategy~(\STAIRm) in Section \ref{sec:training}. All these model are trained with the same datasets described in Section \ref{ssec:datasets}. 

In the experiment, we train a CLIP model and a STAIR model for comparison. For both models, we adopt transformer~\cite{vaswani2017attention} as the backbone with a modified CLIP-B/16~\cite{clip} configurations. The text encoder is a 12-layer Transformer with 512 hidden dimensions and 8 attention heads. The text input is tokenized by BERT WordPiece tokenizer~\cite{kenton2019bert} with 30,522 vocabularies. The max input sequence length is set to 76. The image encoder is a 12-layer transformer with 12 attention heads and 768 hidden dimension sizes. 

The CLIP model is trained for 600K steps using a LAMB optimizer~\cite{You2020Lamb} with a learning rate of $1e^{-3}$ and a weight decay ratio of $1e^{-2}$. The learning rate is first warmed up till 10k steps and then linear decay to zero. \STAIRm model takes 300K, 300K, and 600K steps for each of the stage, where each stage use the same configuration as above. To avoid catastrophic forgetting \cite{mccloskey1989catastrophic,french1999catastrophic}, a smaller max learning rate of $1e^{-4}$ is adopted in stage 3. FLOPs regularization weights are set to \(\lambda_1 = \lambda_2 = 1e^{-3}\) by default with a quadraitc growth following \cite{formal2021spladev2}. All models are trained using a global batch size of 16,384 if not specified.

\subsection{Zero-Shot Text Image Retrieval}

Table \ref{tab:zero-shot retrieval} shows the recall@K~(K=1, 5, 10) performance of image/text retrieval tasks on Flickr30K \cite{plummer2015flickr30k} and COCO-5K \cite{coco_captions}. The metrics are reported with a prompt of ``a photo of '' added before the original caption following \citet{clip}. We observe great performance improvement of STAIR models over the CLIP baseline, \ie 4.9\% and 4.3\% improvement on COCO-5K text$\rightarrow$image and image$\rightarrow$text retrieval recall@1 respectively. A similar improvement is observed on Flickr30K.

\subsection{Zero-Shot Visual Classification}
In Table \ref{tab:zero-shot classification}, we evaluate the models on the zero-shot image classification on 9 datasets from \citet{clip}. We report the top-1 accuracy and employ the same prompts set from \citet{clip}. Results suggest that \STAIRm performs better or as competitive comparing CLIP on most datasets. In particular, \STAIRm shows significantly better performance on SVHN classification that requires an exact match, thanks to its token grounding capability. On the other side, we notice that STAIR also suffers on specialized out-of-distribution tasks, such as Eurosat \cite{helber2019eurosat} and RESISC45 \cite{cheng2017remote}, similar to CLIP as mentioned by \citet{clip}.

\subsection{Linear Probe of Visual Classification}
We also compare the linear probe performance using the same 9 datasets as zero-shot classification. We observe that the STAIR model shows more prominent results comparing the CLIP, even on Eurosat and RESISC45 datasets, where it shows weaker performance in zero-shot. It is because the STAIR model can benefit from its embedding dimensionality of 30,522, which is much larger than 512 for CLIP. Moreover, with sparsity enforced in STAIR, there is no extra cost on storage and computation compared to the dense embeddings (More discussion in Section \ref{sssec:ablation_sparsity}).

\section{Interpretability}
\label{sec:interpret}

%auto-ignore
\begin{table}[t]
  \caption{{\bf Improved interpretability of the STAIR model.} We report the top-K accuracy (\%) of the label among all of the vocabulary in the tokenizer on the ImageNet, CIFAR-100, and CalTech-101 datasets.}
  \vskip 0.15in
\small
    \begin{subtable}{0.5\textwidth}
      \centering
      \begin{tabular}{lrrrr}
        \toprule
        \bf{ImageNet\space\quad} & Top-1 & Top-10 & Top-50 & Top-100\\
        \midrule
        \STAIRm & 32.9 & 69.0 & 83.8 & 87.7 \\
        CLIP & 13.7 & 34.3 & 47.0 & 51.9 \\
        \bottomrule
      \end{tabular}
    \end{subtable}
        \hfill
    \begin{subtable}{0.5\textwidth}
      \centering
        \begin{tabular}{lrrrr}
        \toprule
        \bf{CIFAR-100\space\space} & Top-1 & Top-10 & Top-50 & Top-100\\
        \midrule
        \STAIRm & 10.3 & 56.8 & 75.4 & 80.7 \\
        CLIP & 8.0 & 28.9 & 44.7 & 50.5 \\
        \bottomrule
        \end{tabular}
    \end{subtable}
        \hfill
    \begin{subtable}{0.5\textwidth}
      \centering
        \begin{tabular}{lrrrr}
        \toprule
        \bf{CalTech-101} & Top-1 & Top-10 & Top-50 & Top-100\\
        \midrule
        \STAIRm & 29.3 & 45.4 & 56.0 & 64.8 \\
        % \STAIRs & 0.0\% & 0.0\% & 0.1\% & 0.2\% \\
        CLIP & 8.1 & 24.2 & 38.9 & 43.8 \\
        \bottomrule
        \end{tabular}
    \end{subtable}
  \label{tab:stage_interp}
\end{table}

\begin{figure*}[t]
\centering
\small
     \begin{subfigure}[b]{0.31\textwidth}
         \centering
         \includegraphics[width=\textwidth]{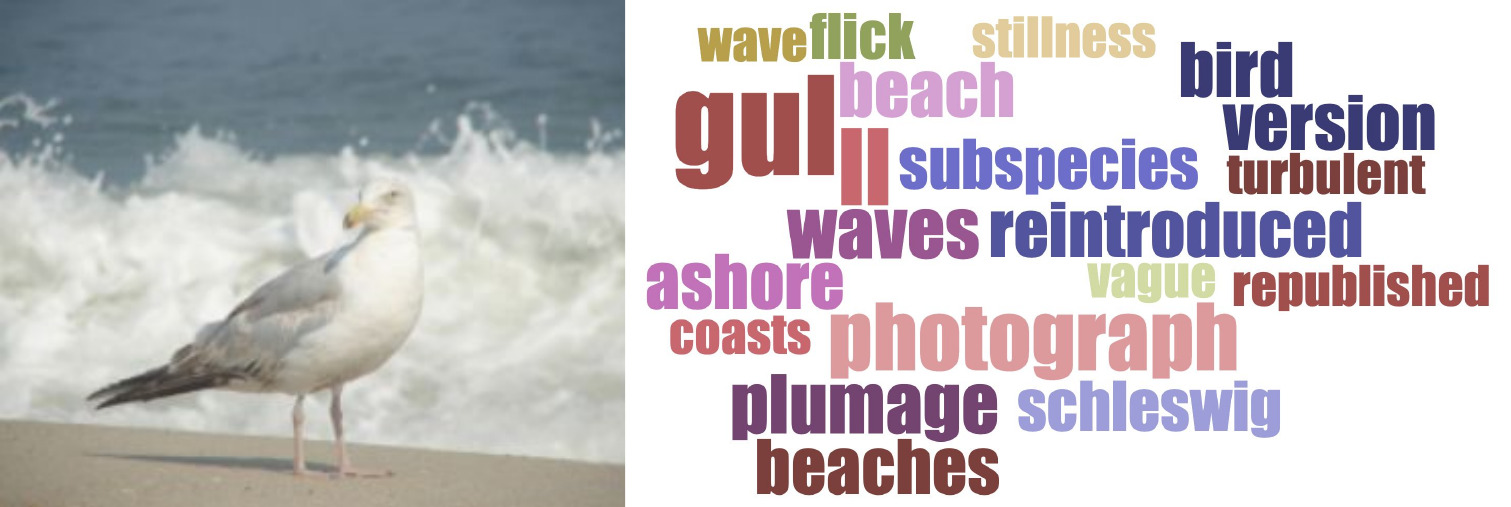}
         \caption{\textbf{Original Caption}:  A seagull standing on the sand of a beach.}
     \end{subfigure}
     \hfill
     \begin{subfigure}[b]{0.31\textwidth}
         \centering
         \includegraphics[width=\textwidth]{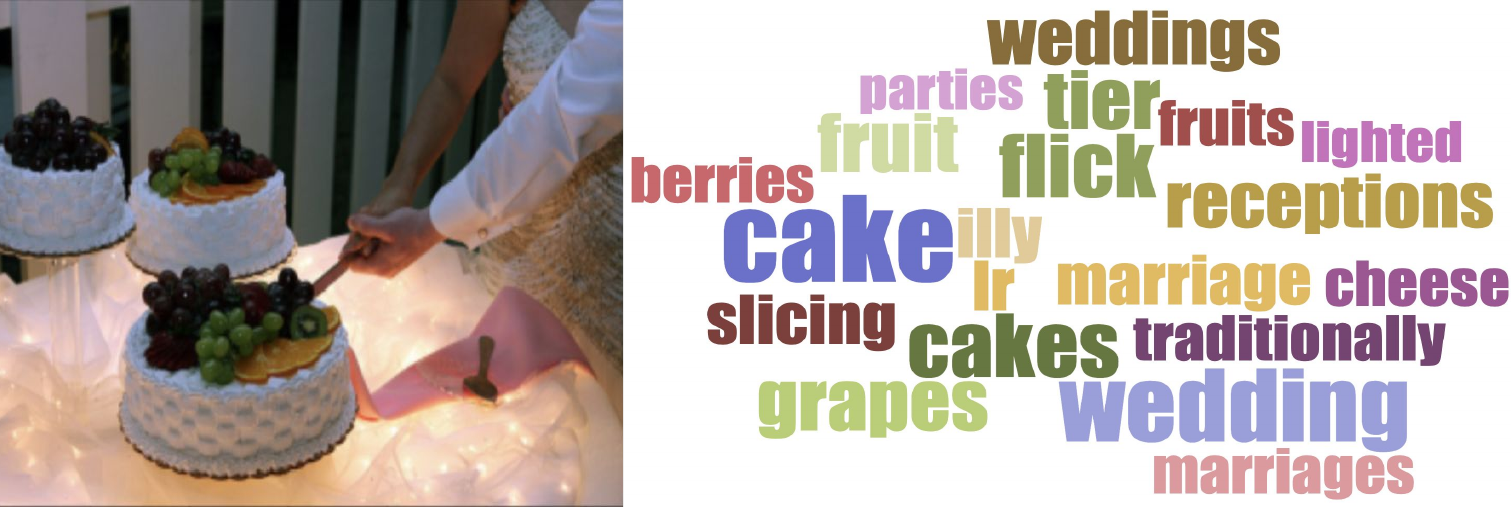}
         \caption{Bride and grooms arms cutting the wedding cake with fruit on top.}
     \end{subfigure}
     \hfill
     \begin{subfigure}[b]{0.31\textwidth}
         \centering
         \includegraphics[width=\textwidth]{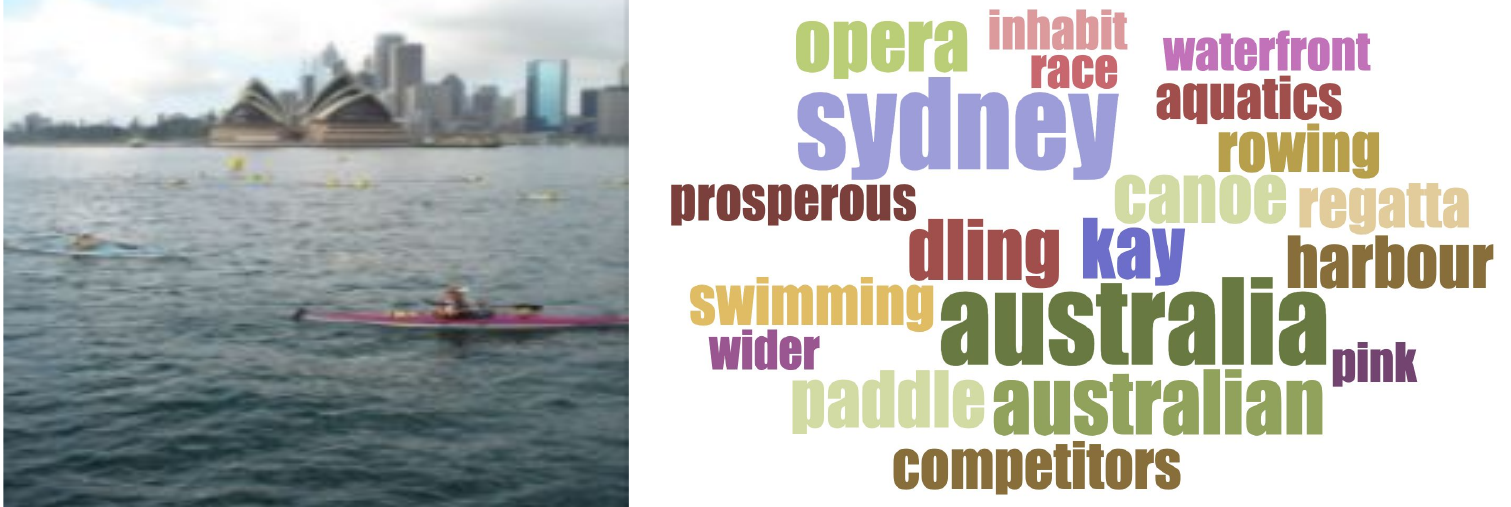}
         \caption{A couple of people on kayak boats in the middle of the ocean.}
         % \caption{A couple of people on kayak boats in the middle...}
     \end{subfigure}
     \hfill
     
     \bigskip
     \begin{subfigure}[b]{0.31\textwidth}
         \centering
         \includegraphics[width=\textwidth]{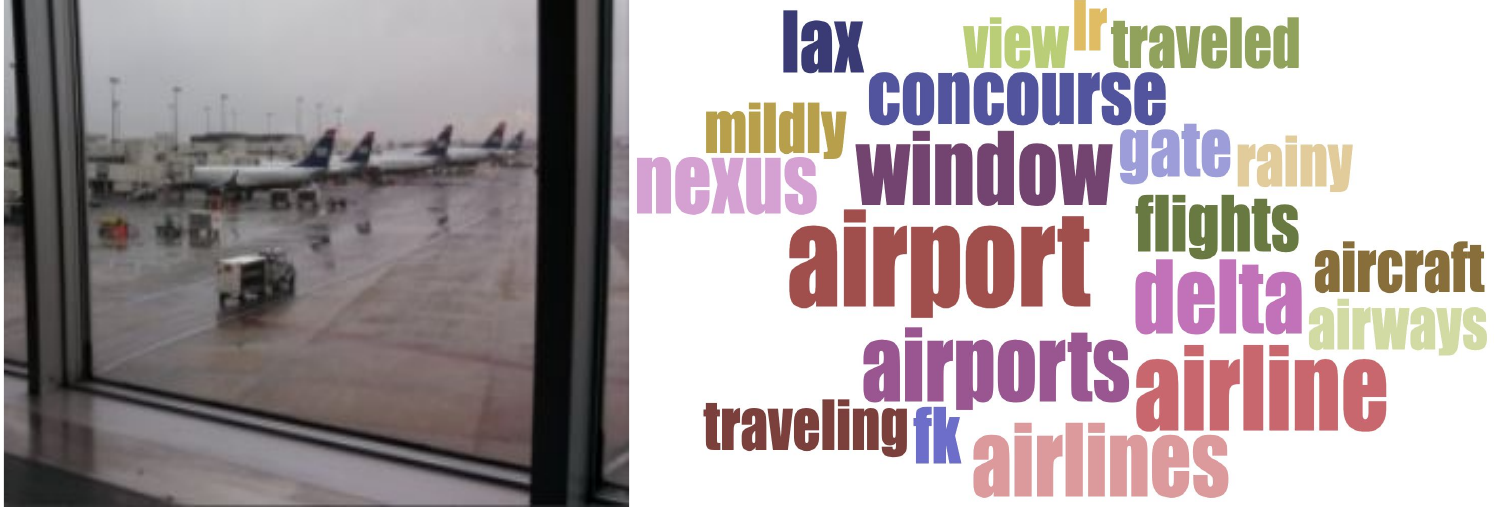}
         \caption{An airport filled with planes sitting on tarmacs.}
     \end{subfigure}
     \hfill
     \begin{subfigure}[b]{0.31\textwidth}
         \centering
         \includegraphics[width=\textwidth]{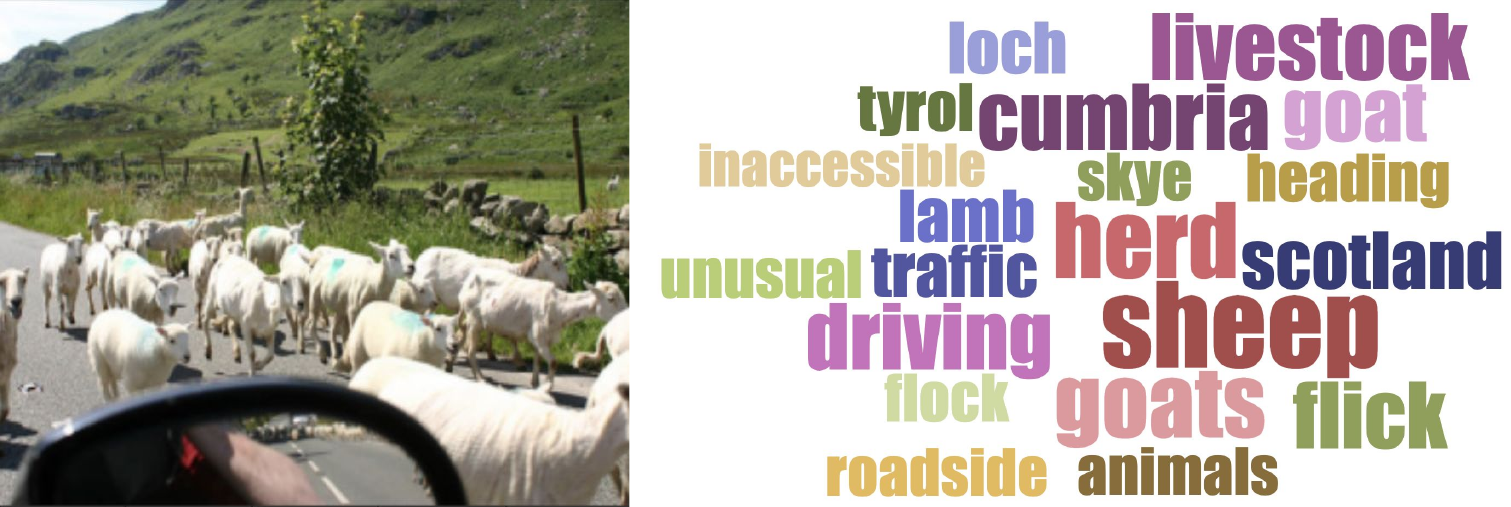}
         \caption{Sheared sheep on roadway taken from vehicle.}
     \end{subfigure}
     \hfill
     \begin{subfigure}[b]{0.31\textwidth}
         \centering
         \includegraphics[width=\textwidth]{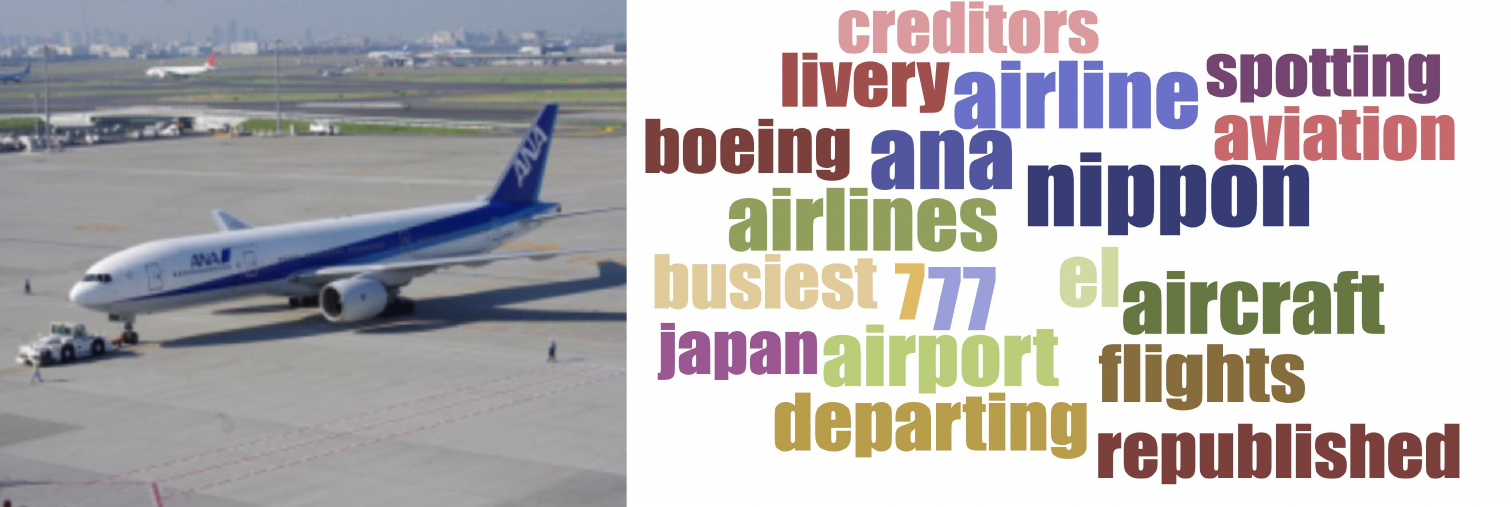}
         \caption{A plane on the runway is being led by a tow cart.}
     \end{subfigure}
     \hfill
     
     \bigskip

    % \hspace{2cm}
     \begin{subfigure}[b]{0.31\textwidth}
         \centering
         \includegraphics[width=\textwidth]{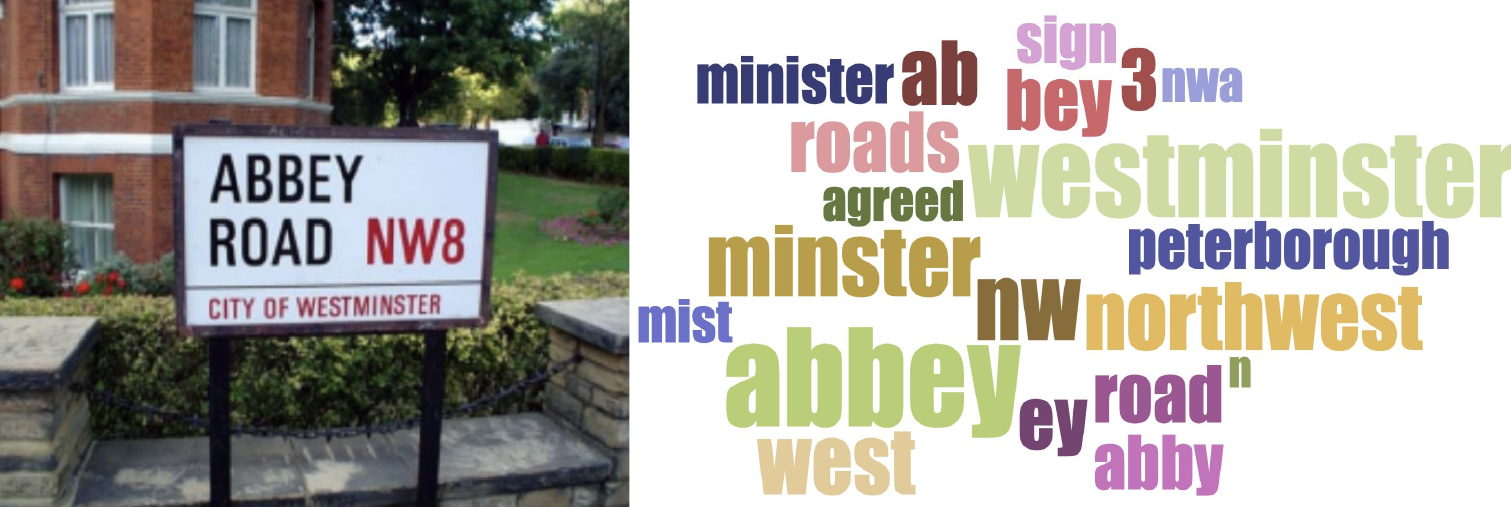}
         \caption{There is a sign in front of a brick house.  \quad \quad \quad \quad\quad \quad}
     \end{subfigure}
     \hspace{0.3cm}
     \hfill
     \begin{subfigure}[b]{0.31\textwidth}
         \centering
         \includegraphics[width=\textwidth]{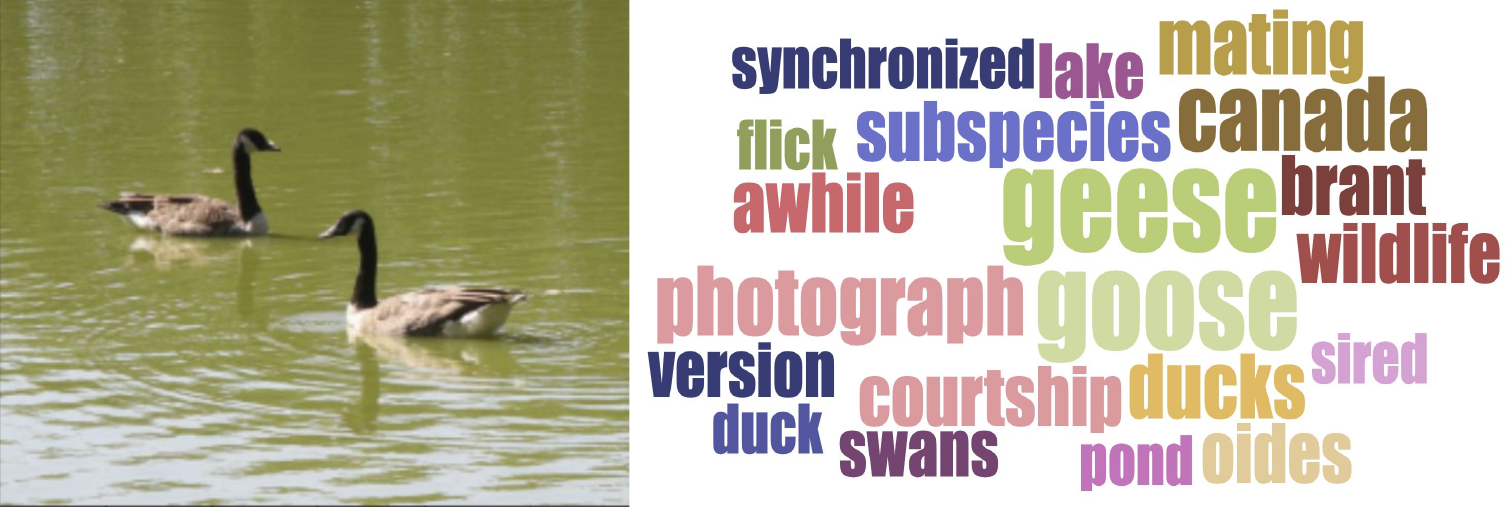}
         \caption{Two ducks floating together on water.  \quad \quad \quad \quad \quad \quad \quad}
     \end{subfigure}
     \hspace{0.3cm}
     \hfill
     \begin{subfigure}[b]{0.31\textwidth}
         \centering
         \includegraphics[width=\textwidth]{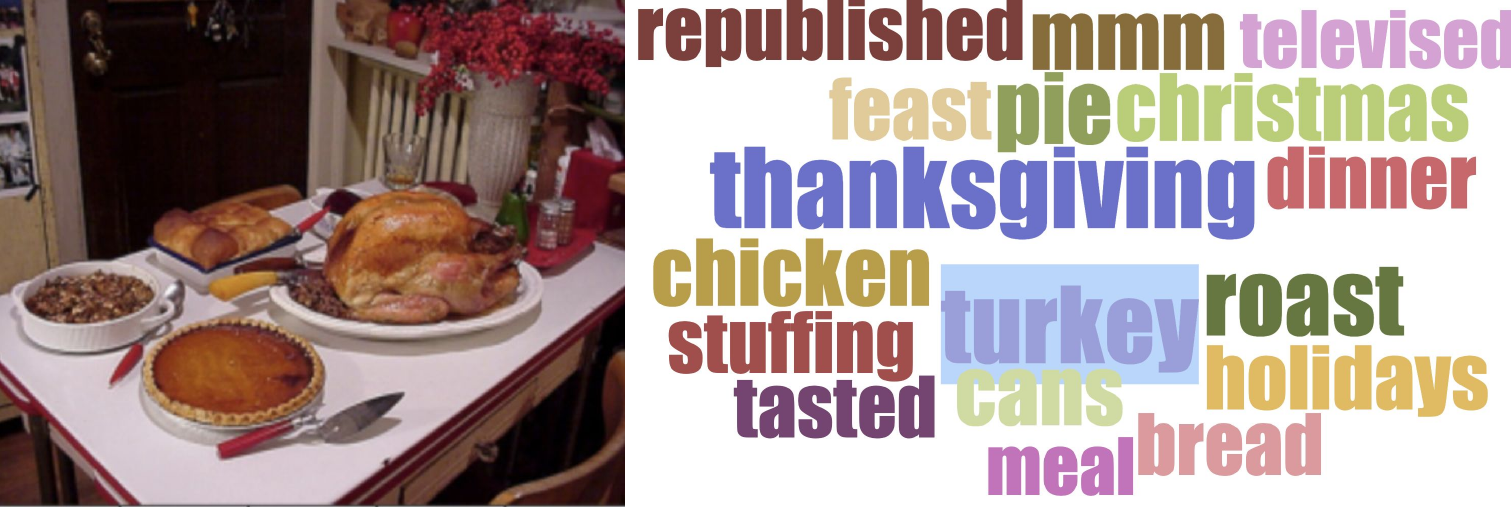}
         \caption{A table set for a traditional Thanksgiving dinner.}
     \end{subfigure} 
     \hfill
     % \vspace{-0.1in}
\caption{{\bf Selected examples of predictions from STAIR model.} Top 20 tokens predicted from an image using \STAIRm where the font size indicates the prediction weight. Note the original caption is shown below the image. Detailed weights are listed in Appendix \ref{sec:image_token_examples_weights}.}
\label{fig:image_token_examples}
\end{figure*}

\begin{figure}[t]
\begin{center}
\includegraphics[width=0.48\textwidth]{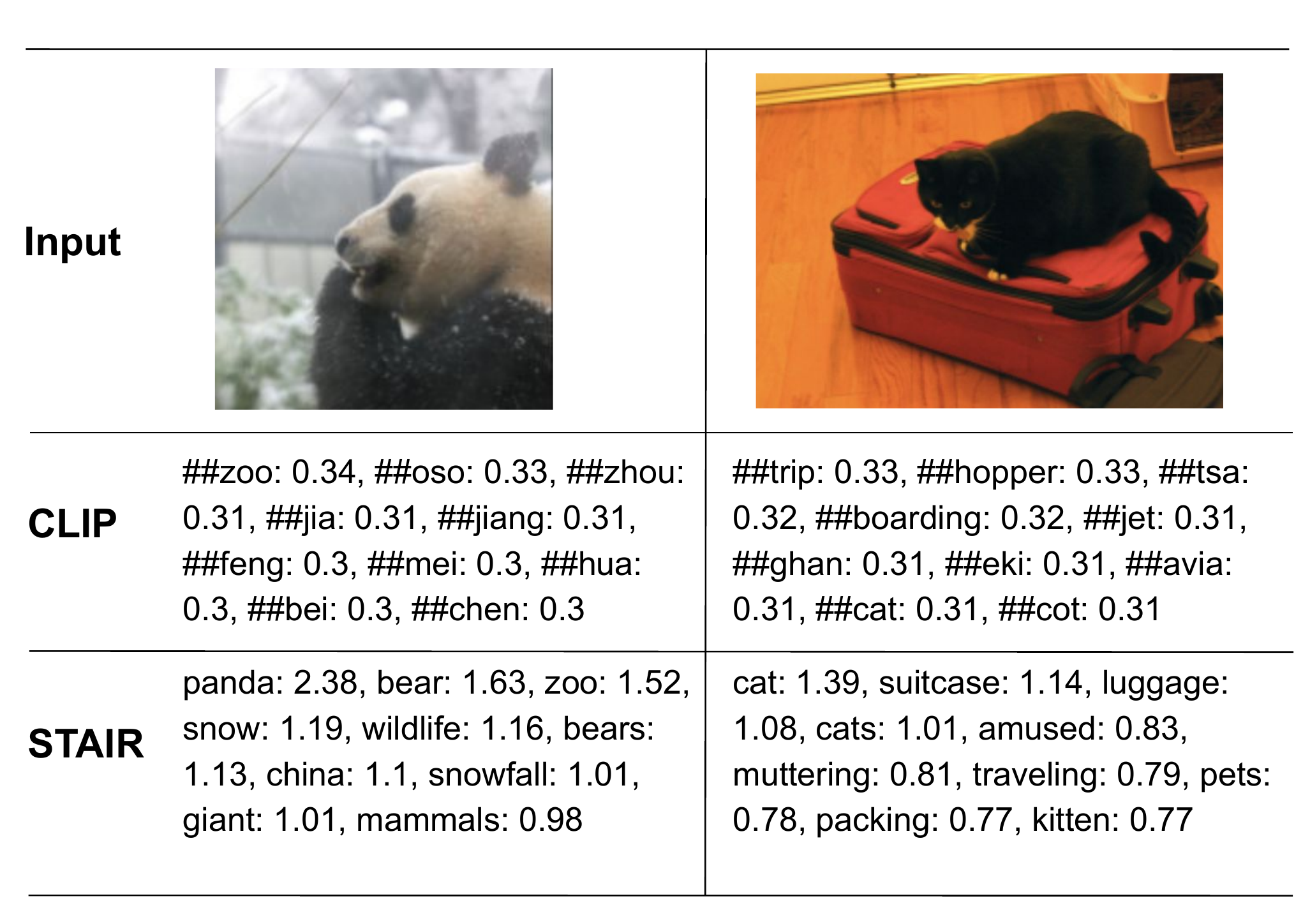}
\end{center}
\vspace{-0.2cm}
\caption{{\bf Selected examples of most relevant (sub-)words in the lexicon space and the similarities from CLIP and STAIR.} 
Tokens from STAIR are more friendly for human to interpret the visual concepts than CLIP. \textit{\#\# indicates subword from the vocab.}}
\label{fig:stair_clip_vs_ms}
\end{figure}

\subsection{Quantatively analysis}

A key difficulty of interpreting a high-dimensional embedding is that its representation residing in the embedding space $R^d$ does not naturally correspond to concepts that humans easily understand. To address the challenge, \citet{kim2018interpretability} proposed to leverage an \textbf{interpretable space} $R^H$ that is spanned by vectors $c$ corresponding to an unknown set of human-interpretable concepts. From this standpoint, the embedding interpretation can be viewed as a mapping of $R^d\rightarrow R^H$. 

For image-text dual encoder models, a good interpretable space to connect multimodalities is a lexicon space that is spanned by basis vectors $c$ of text embeddings representing human understandable tokens, words, phrases, and/or sentences. The interpretation then can be given by the similarity between the embedding and each basis vector, $Sim(\cdot, c)$. The lexicon space is crucial for comparing the interpretation between different models. Suppose we have the embedding from a dog image, without a dog concept existing in the lexicon space, it is infeasible to understand from the image embedding itself. 

Zero-shot image classification is a restricted form of functionally-grounded interpretability evaluation \cite{doshi2017towards} as its interpretable space is predefined by its classes and dataset specific. In practice, the interpretable space can be both lack of human labels and unlimited \cite{ghorbani2019towards}. To lift the constraint, we expand our interpretable space as the vocabulary of the BERT WordPiece Tokenizer~\cite{kenton2019bert} ($n=30,522$) to approximate a lexicon space covering any human-interpretable concepts. Note that, under this definition, the embedding space becomes the same as the interpretable space for STAIR and the interpretation mapping reduces to an identity function. Similar to zero-shot image classification, if an image embedding lies closer to its ground-truth class in the lexicon space, we consider it as easier to interpret. This task is usually more challenging than the original zero-shot classification because the candidate class now becomes the entire vocabulary, which is much more than the predefined classes.

We compare the interpretability of CLIP and STAIR on three datasets, ImageNet, CIFAR-100, and Caltech-101. The Top-K accuracy is used to quantitatively measure the model’s interpretability. In particular, if a ground-truth class is tokenized into separated sub-words $c_k$ in the vocabulary, we take the max similarity over the sub-words $\max_{k}(Sim(\cdot, c^k))$ as the final similarity. As shown in Table \ref{tab:stage_interp}, it is clear that STAIR provides significantly better interpretability than the CLIP model in the interpretable space.

\subsection{Qualitatively analysis}
\label{sec:grounding}

Here we qualitatively examine that the sparse embedding from the STAIR model is more interpretable compared to CLIP. In Figure \ref{fig:stair_clip_vs_ms}, we report the top $Sim(\cdot, c^k)$ (sub-)words in the interpretable space defined by the BERT WordPiece Tokenizer for each image. We see that STAIR is better at capturing visual concepts that humans can easily understand than CLIP, which is consistent with our quantitative analysis. We also observe that the top tokens from CLIP tend to have similar matching scores while STAIR avoids the problem by adopting eq.~\eqref{eq:log_max_pooling} in the Token Projection Head. Figure \ref{fig:image_token_examples} shows more examples of (sub-)words with the highest weights from STAIR embeddings given each image. The results suggest that STAIR is capable of grounding images to tokens that are semantically related. For example, it can infer ``wedding'' from the picture of bride and groom cutting cake. In practice, this grounding and interpretability ability is very useful for debugging and understanding model’s behavior. For example, we observe the model favors activating ``https'' token in many predicted embeddings. We find that the bias is mainly caused by a large portion of web-mined content in our training data, where ``https'' occurs in many of the associated texts.

\section{Analysis and Discussion}
\label{sec:analysis}
%auto-ignore
\begin{figure}[t]
    \centering
    \vspace{-0.2cm}
    \includegraphics[width=0.48\textwidth]{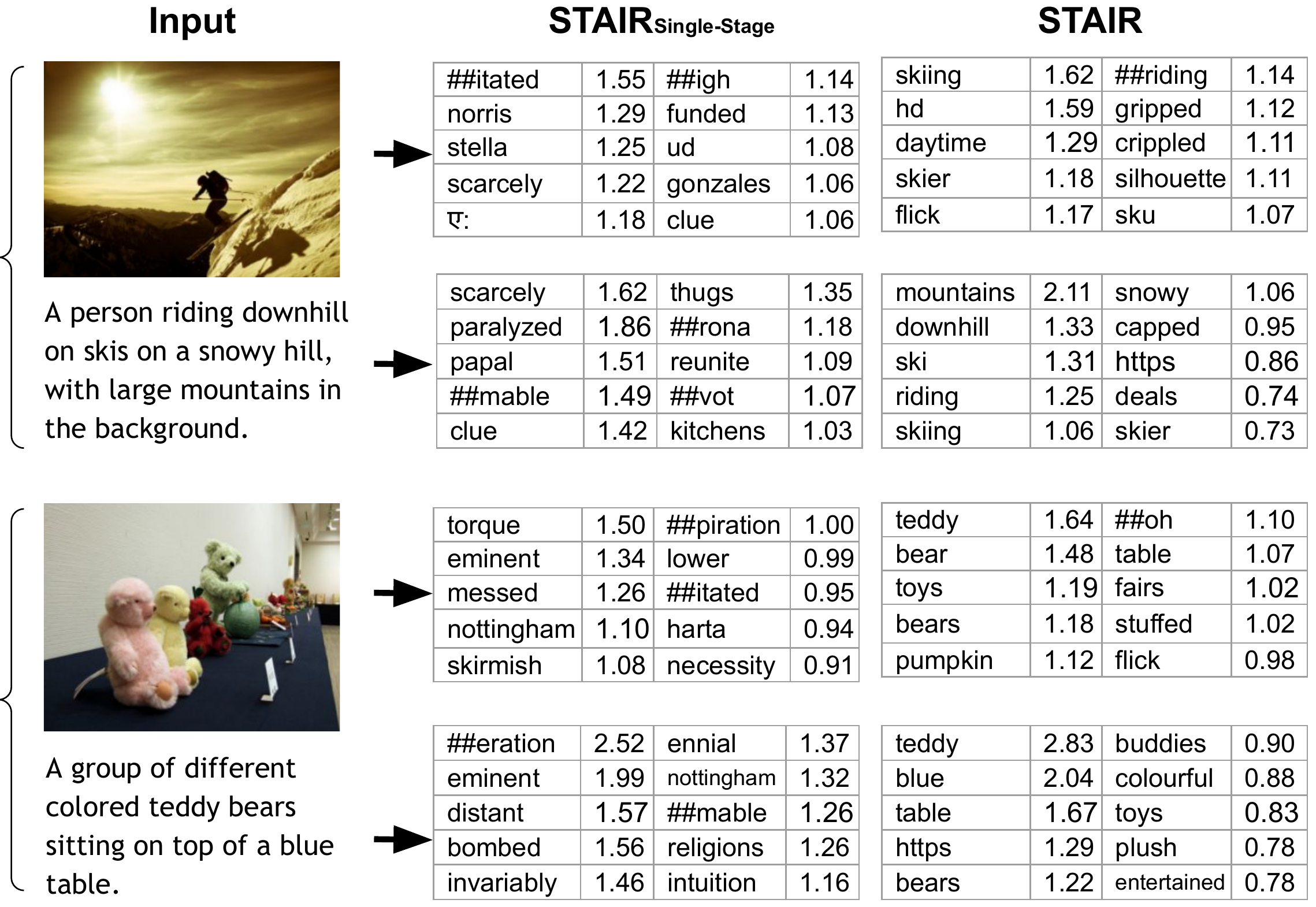}
    \caption{{\bf Selected examples of top predictions and their weights from STAIR models.} The predicted tokens from \STAIRs are not connected the actual meaning of inputs, while \STAIRm predictions from multi-stage training are grounded to input semantics. \textit{\#\# indicates subword from the vocab.}}
    \label{fig:token_grouding}
\end{figure}

\subsection{Necessity of Multi-stage training}

Multi-stage training strategy is crucial to guarantee that STAIR embedding grounds to meaningful tokens. To demonstrate its necessity, we train a single-stage model, denoted as \STAIRs for comparison. Figure~\ref{fig:token_grouding} demonstrates the predictions from \STAIRs and multi-stage \STAIRm models separately. We observed that \STAIRs tends to redefine the semantic meaning of tokens and reuse them to match images and text. For example, \STAIRs does not interpret tokens ``eminent'' and ``nottingham'' as famous and locations but redefines them as teddy bear topics, as they are always the top two predicted tokens for teddy bear images. Similarly, ``clue'' are re-interpreted as the concept of skiing. Although we can guess and infer the new semantic meaning of each token through reverse engineering, the redefined semantic meaning is far from its original one. This makes it hard for humans to interpret the predicted embeddings. In contrast, by adding multi-stage training, tokens are grounded to their original meaning and the embedding is human-readable~\footnote{Interestingly, \STAIRs model can still achieve comparable performance as the multi-stage training on various retrieval and classification tasks, better than CLIP. We hypothesize that it benefits from its sparse embedding, whose dimensionality is much larger than CLIP. More details of the \STAIRs training and quantitative results can be found in Appendix \ref{sec:single_stage}.}.

\subsection{Ablation on Multi-stage training}
\begin{figure}[t]
     \centering
     \includegraphics[width=0.48\textwidth]{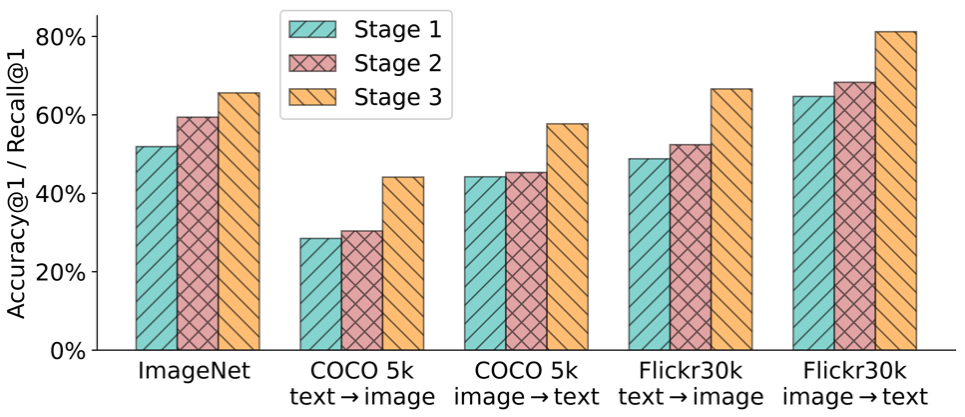}
     \caption{{\bf Ablation on multi-stage training strategy.} Stage 1 has laid a reasonable foundation for zero-shot performance. Stage 2 helps more on ImageNet than Stage 1. Stage 3 further improves both on ImageNet and COCO/Flickr30K.}
     \label{fig:stage_ablation}
 \end{figure}

In this section, we qualitatively study the effect of the multi-stage training strategy on zero-shot transfer. Figure \ref{fig:stage_ablation} shows the top-1 accuracy of ImageNet classification and recall@1 on COCO-5K and Flickr30K image-text retrieval for each stage separately. We observe that stage 1 can already achieve reasonable performance even though the activation of text token is restricted while stage 2 helps more on the classification task. Furthermore, stage 3 greatly boosts the text-image matching ability of the text and image encoders together via contrastive learning, which is reflected in all metrics.

\subsection{Ablation on Sparsity}
\label{sssec:ablation_sparsity}
\begin{figure}[t]
    \centering
     \begin{subfigure}[b]{0.48\textwidth}
         \centering
         \includegraphics[width=\textwidth]{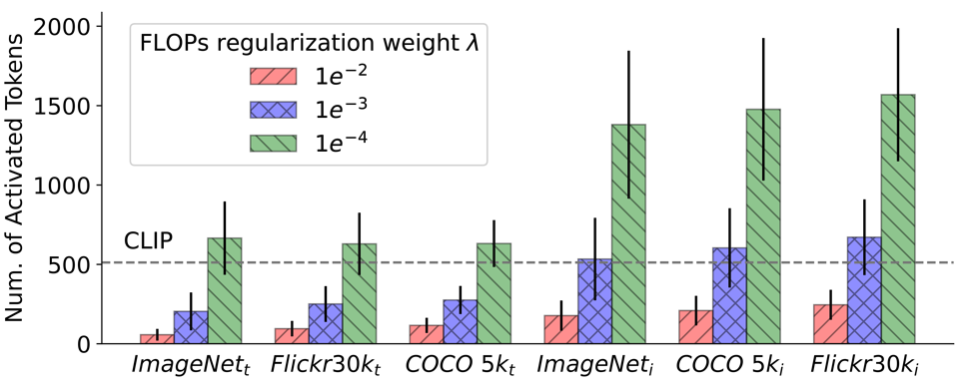}
         \caption{{\bf Number of tokens activated.} Subscript $t$, $i$ indicates the results are from text or image. The horizontal line represents the dense CLIP baseline with embedding size of 512. The effective text embedding size from STAIR is significantly lower than CLIP, which helps speed up the retrieval.} 
     \end{subfigure} 
      \begin{subfigure}[b]{0.48\textwidth}
         \centering
         \includegraphics[width=\textwidth]{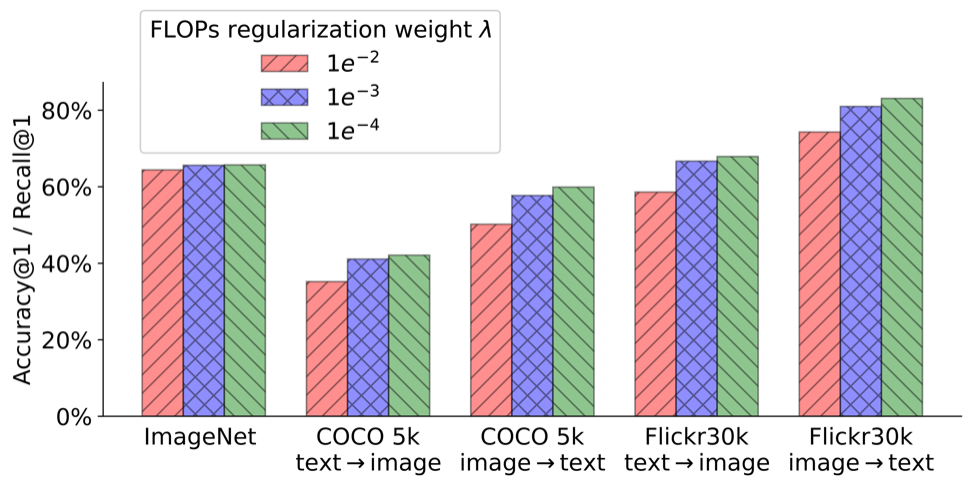}
         \caption{{\bf Performance on zero-shot transfer.} More sparsity in embedding lowers the zero-shot transfer performance.}
     \end{subfigure}    
\caption{{\bf Ablation on FLOPs regularization weight $\lambda$.} }
\label{fig:sparsity_ablation}
    \vspace{-0.5cm}
\end{figure}

The embedding sparsity is essential to guarantee the efficiency of similarity computation and retrieval speed. In STAIR, the strength is controlled by the FLOPs regularization weights. To study its impact, we train three \STAIRm models with regularization weights $\lambda=\lambda_1=\lambda_2\in\{1e^{-2}, 1e^{-3}, 1e^{-4}\}$. We check their corresponding text and image embedding sparsity, \ie the number of tokens being activated in the predictions, as well as their zero-shot transfer performance. The dataset we use is ImageNet, COCO-5k, and Flickr30k, and the results are summarized in Figure \ref{fig:sparsity_ablation}.

 The results suggest that the regularization weights have a positive impact on encouraging sparsity. Importantly, the effective number of tokens from STAIR is significantly lower than the dense embedding dimension of 512 used in CLIP for the text embedding and comparable for the image embeddings in all 3 settings. Since the complexity of sparse embeddings dot product is linear to the smaller number of non-zero units in two embeddings, STAIR models are more efficient in conducting similarity computation during retrieval compared to CLIP. Moreover, we observe that more tokens are activated in the image embeddings than in the text embeddings. One explanation is that the image semantics is usually broader and more general while the text meaning is more specific as reflected in the sparsity of embeddings. We also notice that when $\lambda$ is large, text embedding tends to be more sparse when the text inputs are shorter. On the other hand, the regularization weights show a negative impact on zero-shot performance. In particular, the trend is more noticeable in retrieval tasks than in image classification.

\section{Text Encoder Free Applications}
%auto-ignore
The token grounding capability also gives \STAIRm model the possibility to tackle existing tasks in a more efficient way. We illustrate two potential applications: \textit{1) text encoder-free retrieval}, and \textit{2) text encoder-free image-text localization}.

\begin{figure}[t]
    \centering
    \vspace{+0.2cm}
    \includegraphics[width=0.48\textwidth]{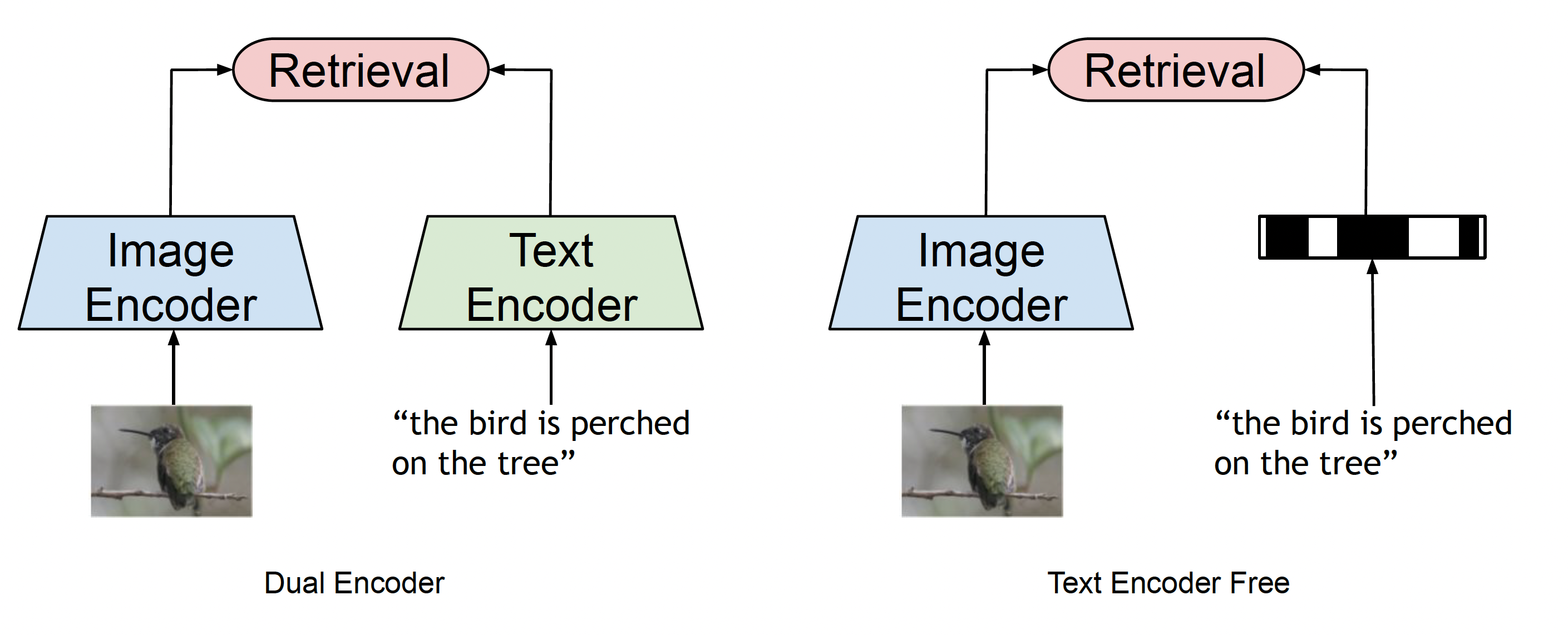}
    \caption{{\bf Dual encoder vs text encoder free architecture.} Text-encoder-free architecture contains only image encoder and uses $\textsc{mask}$ as text embedding.}
    \label{fig:text_encoder_free}
\end{figure}

\begin{table}[t]
\caption{{\bf Zero-shot of \STAIRm with and without text encoder inference.} Reporting recall@K on Flickr30K and COCO, and top-1 accuracy (\%) on ImageNet.}
\vskip 0.15in
\small
    \centering
    \begin{tabular}{l|ccccc}
    \toprule
    & \multicolumn{1}{c}{\bf ImageNet} & \multicolumn{2}{c}{\bf COCO 5K} & \multicolumn{2}{c}{\bf Flickr30K}\\
    & \multirow{2}*{Acc@1} & T2I & I2T & T2I & I2T \\
    & &R@1 & R@1 & R@1 & R@1 \\
    \midrule
    \STAIRm & 65.6 & 41.1 & 57.7 & 66.6 & 81.2 \\
    STAIR\textsubscript{\textsc{image}} & 50.0 & 21.0 & 27.2 & 40.4 & 47.0  \\
    \bottomrule
    \end{tabular}

    \label{tab:text_encoder_free}
    \vspace{-0.1cm}
\end{table}

\subsection{Text Encoder Free Image-Text Retrieval}
Another peculiar advantage of \STAIRm is that it enables the possibility of a text-encoder-free retrieval system. We compare the dual-encoder and text-encoder-free architectures in Figure \ref{fig:text_encoder_free}. More concretely, we use the image encoder of \STAIRm to generate the sparse image embeddings while directly converting texts into $\textsc{mask}$ in the vocabulary space after tokenization as the sparse text embeddings. In contrast to the dual encoders that require inference of both image and text encoder, this architecture only needs the former part, which makes it friendly for applications with restricted latency requirements. 
On the other hand, it is still different from the retrieval system built with fixed taxonomy using an inverted index because it is capable of taking any free text inputs.

In Table \ref{tab:text_encoder_free}, we summarize the zero-shot performance of text-encoder-free \STAIRm compared with the original STAIR. Although it obviously underperforms the dual encoder baseline, the results are still encouraging given its potential. We observe that text-encoder-free \STAIRm shows relatively stronger performance in ImageNet classification compared to text/image retrieval tasks. This is due to the fact that ImageNet classes are more concise in general. In contrast, captions from COCO-5k and Flickr30k usually contain more stop words while they are treated of equivalent importance as the semantically meaningful tokens by $\textsc{mask}$, which indicates large headroom in improving the text-encoder-free performance and we leave it as feature work.

\begin{figure}[t]
    \centering
    \includegraphics[width=.48\textwidth]{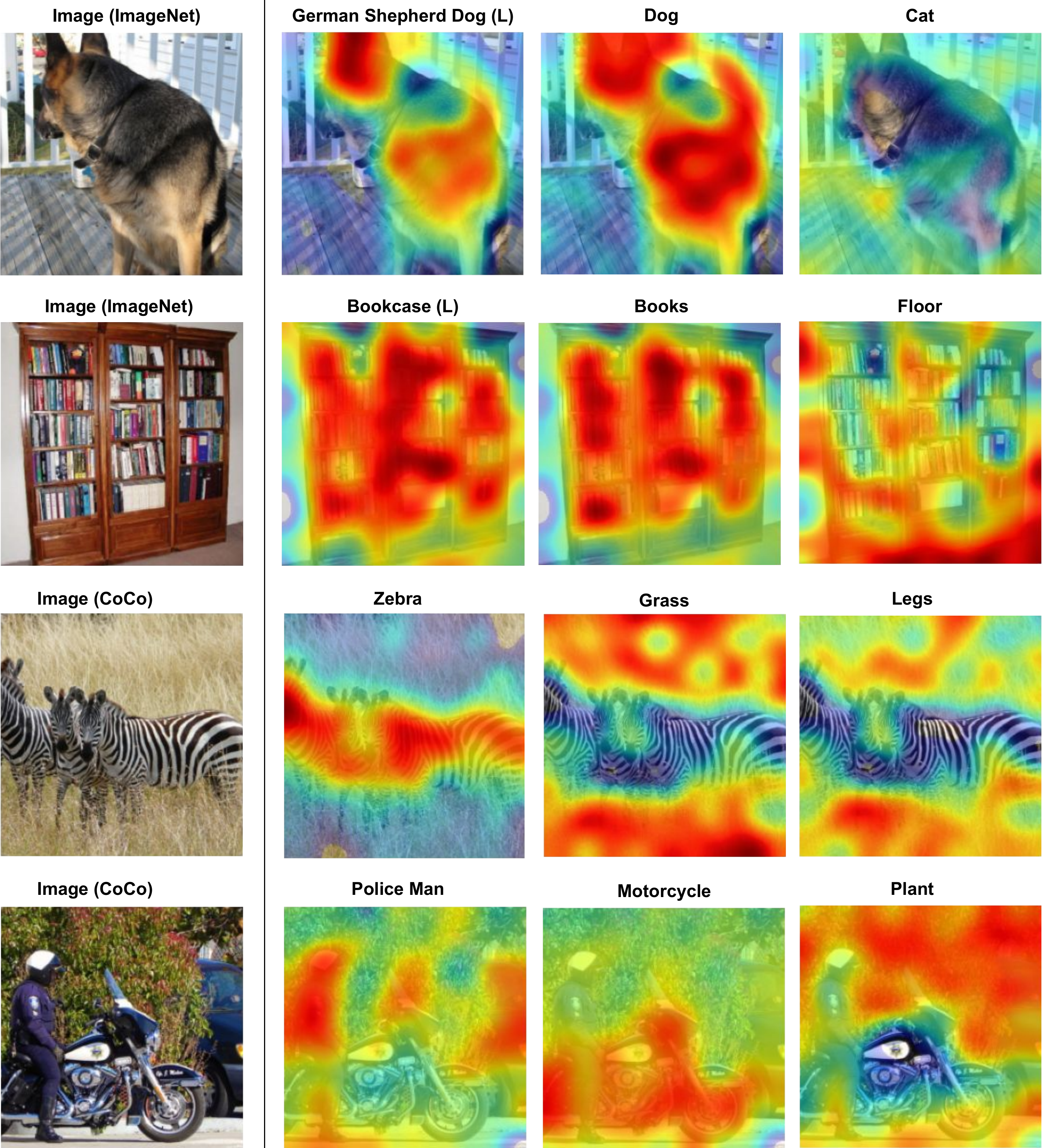}
    \caption{{\bf Qualitative examples of text inference free localization in \STAIRm}. Left side shows the original images from ImageNet or COCO 5K. Right side shows the queries and their activation heatmaps. \textbf{(L)} indicates the query is the original ImageNet label of the image. (The heatmap colors are normalized for better visualization, not representing real values.)}
    \label{fig:localize}
\end{figure}

\subsection{Text Encoder Free Localization}
\STAIRm is able to localize the image regions given an arbitrary query (e.g. an object) from the vocabulary without any inference from the text side~\footnote{If a query is not in the vocabulary or has multiple tokens, we can still do that by tokenizing the query into in-vocab tokens and take the average.}.

Recall that the image encoder uses a vision transformer that splits the original image into grids. In Equation \ref{eq:gmap} the mapping function $p(\cdot)$ projects each token/grid representation to the vocabulary space, where each dimension represents the activation of the corresponding token from the vocabulary. Inspired by this, we can find the regions that are correlated to the query by simply looking at the activation score of given input tokens at each grid. Figure~\ref{fig:localize} shows examples of images given arbitrary text queries and their activation heatmaps. In the first example we show the activation map for queries ``German Shepherd Dog'', ``Dog'', and ``Cat''. Both ``German Shepherd Dog'' and ``Dog'' are well aligned with the actual dog in the image. In contrast, the activations  of the query ``Cat'' are across the entire image. We also notice the activation heatmap of ``Dog'' is aligned better compared to the activation heatmap  of ``German Shepherd Dog''. It is because the multi-token query is decomposed into multiple tokens from the pre-defined vocabulary, and tokens ``German'' and ``Shepherd'' usually represent other concepts rather than a dog. A better-tuned vocabulary, e.g. including the ``German Shepherd Dog'' as a token, will help solve this problem.

We hypothesize that \STAIRm models get the localization capability because of the max pooling operation. According to \cite{kanchana2022}, simply changing the image and text encoder poolers to max pooling can greatly improve the localization and segmentation capability of a CLIP model. The difference is that CLIP model still needs the encode the text to align with the patch representations from the image, while \STAIRm doesn't require any inference from the text side.
The findings show STAIR model has great potential for localization, segmentation, and open vocabulary detection tasks. We leave it as future work.

\section{Related Work}
\label{sec:related}
%auto-ignore
\paragraph{Image and text retrieval} There are two categories of approach for image-text retrieval: dual-encoder approach and cross-encoder approach. Both approaches encode images and text as dense embeddings. Contrastive loss is calculated from the dense embeddings to optimize the model. Frome \etal proposed DeVISE~\cite{frome2013devise} as the first dual-encoder approach for image-text retrieval. The model contains an image encoder and a text encoder that encodes the image and text input as dense embeddings. Cosine distance is adopted to measure the similarity of image and text pairs. With the advent of transformers, \citet{clip} proposed CLIP that leverages large-scale pretraining datasets and established new state-of-the-art across multiple benchmarks. Dual encoder usually achieves faster retrieval speed, as the embeddings can be pre-computed and cached. Finetuning the pre-trained visual transformer model ~\cite{dosovitskiy2020image} and language model ~\cite{kenton2019bert} can further boost its performance. Contrary to dual-encoder, the cross-encoder approach uses a single encoder to handle both image and text inputs. UNITER~\cite{chen2020uniter} concatenates the image and text as a single input sequence and employs a multi-layer transformer model to encode them jointly. Compared to the dual-encoder, the cross-encoder is slower at inference but achieves better performance on the benchmarks, as it is capable of capturing the image and text alignment at multiple granularities. 

STAIR follows the dual-encoder approach. However, instead of encoding the image and text as dense embeddings, it encodes them as \textit{sparse embedding}. In particular, the sparse embedding from STAIR can be easily human-interpretable and is shown to achieve better results compared to dense embeddings.

\paragraph{Document retrieval via sparse embedding} Retrieving documents using sparse embedding is a popular approach in information retrieval ~\cite{dai2019context,bai2020sparterm,formal2021splade,formal2021spladev2}. As the number of documents increases, the sparse embedding can be hashed and allows fast retrieval using an inverted index system ~\cite{formal2021splade,formal2021spladev2}. Our approach is largely inspired by SPLADE~\cite{formal2021splade} which encodes the text query and documents as sparse embedding and uses an inverted index system to conduct text retrieval. Unlike SPLADE, our approach handles retrieval across modalities, which imposes unique challenges. Due to the semantic gap between image and text, designing a joint sparse embedding space for both modalities is a non-trivial task. Furthermore, grounding the image and text to meaningful tokens in the joint embedding space is challenging. In STAIR, we proposed a streamlined approach to enable fast retrieval speed, interpretation, and high retrieval accuracies.

\section{Conclusion}
In this paper, we proposed the sparse text and image representation approach (STAIR) that encodes image and text inputs into sparse embeddings in a sparse token space and multi-stage training strategy to guarantee that the embeddings grounds to meaningful tokens. We compare the STAIR model with CLIP model and results show that STAIR model can significantly outperform on image-text retrieval tasks, and also achieved better performance on various zero-shot and linear probing classification tasks. Moreover, we conduct quantitative and qualitative analysis to demonstrate that our sparse embedding is easy for human to interpret comparing to dense embedding.  

\section*{Acknowledgements}
We thank our teammates from Visual Intelligence, foundation model, and other Apple groups for their feedback and suggestion. Sepcial thanks goes to Navdep Jaitly, Brandon McKinzie, Xianzhi Du for the early review of the paper, Floris Weers, Vaishaal Shankar for linear probing experiments setup and Cybertron team for infrastructure supports.

\bibliography{egbib}
\bibliographystyle{icml2023}

%%%%%%%%%%%%%%%%%%%%%%%%%%%%%%%%%%%%%%%%%%%%%%%%%%%%%%%%%%%%%%%%%%%%%%%%%%%%%%%
%%%%%%%%%%%%%%%%%%%%%%%%%%%%%%%%%%%%%%%%%%%%%%%%%%%%%%%%%%%%%%%%%%%%%%%%%%%%%%%
% APPENDIX
%%%%%%%%%%%%%%%%%%%%%%%%%%%%%%%%%%%%%%%%%%%%%%%%%%%%%%%%%%%%%%%%%%%%%%%%%%%%%%%
%%%%%%%%%%%%%%%%%%%%%%%%%%%%%%%%%%%%%%%%%%%%%%%%%%%%%%%%%%%%%%%%%%%%%%%%%%%%%%%
\newpage
\appendix
\onecolumn

\section{Single-stage training vs Multi-stage Training}
\label{sec:single_stage}
We train a STAIR model without using the multi-stage training strategy described in Section \ref{sec:training}, denoted as \STAIRs.
 The \STAIRs model shares the same model architecture as shown in Figure \ref{fig:stair_diagram} with the multi-stage STAIR model and is trained with the same training configurations as the CLIP model as described in Section \ref{ssec:config}. The FLOPs regularization weights are set to the same value as the \STAIRm model, \ie $\lambda_1 = \lambda_2 = 1e^{-3}$.

Table \ref{tab:single_stage_retrieval} lists the zero-shot image-text retrieval and classification performance of two versions of the \STAIRm model and the baseline CLIP model. Compared to the multi-stage trained model, \STAIRs can achieve similar performance on zero-shot text/image retrieval and classification tasks. They both outperform the CLIP model in most of the metrics. The interpretability, however, is much worse as shown in Table \ref{tab:single_stage_interp}. As illustrated in Section \ref{sec:grounding}, \STAIRs model plays a role as a multi-modal clustering algorithm, which repurposes the words as weighted token centroids. The contrastive objective trains the model to focus on matching the aligned image and text using the token centroids but is inadequate in restricting the predicted tokens to their original human-readable meanings.

\begin{table}[h]
\caption{{\bf Zero-shot text/image retrieval and classification.}  Reporting recall@K on Flickr30K and COCO, and top-1 accuracy (\%) across 9 classification datasets. Bold indicates the best overall performance.}
\vskip 0.15in
%\caption{Recall@K(\%) on Flickr30K and MSCOCO Image-Text Retrieval Tasks (Zero-shot).}
\small
\begin{subtable}{1\textwidth}
  \centering
  \begin{tabular}{l|cccccc|cccccc}
    \toprule
    \multirow{3}{*}{\bf{Retrieval}}
    & \multicolumn{6}{c|}{\bf COCO 5K} & \multicolumn{6}{c}{\bf Flickr30K}\\
    & \multicolumn{3}{c}{text $\rightarrow$ image} & \multicolumn{3}{c|}{image $\rightarrow$ text} & \multicolumn{3}{c}{text $\rightarrow$ image} & \multicolumn{3}{c}{image $\rightarrow$ text}\\
    & R@1 & R@5 & R@10 & R@1  & R@5 & R@10 & R@1 & R@5 & R@10 & R@1 & R@5 & R@10\\
    \midrule
    \rule{-2pt}{8pt}
    CLIP & 36.2	& 62.2 & 72.2 &	53.4 & 78.3 & 85.6 & 63.0 &	86.7 & 92.5 & 79.6 & 95.5 &	98.1 \\
    \STAIRs & 40.5 & 64.9 & {74.9} & {57.7} & {80.3} & {87.4} & \bf{66.7} & {88.6} & \bf{93.7} & {81.0} & \bf{96.7} & \bf{98.5} \\
    \STAIRm & \bf{41.1} & \bf{65.4} & \bf{75.0} & \bf{57.7} & \bf{80.5} & \bf{87.3} & 66.6 & \bf{88.7} & {93.5} & \bf{81.2} & 96.1 & 98.4 \\
    % \bottomrule
  \end{tabular}
\end{subtable}
        %\hfill
\begin{subtable}{1\textwidth}
\small
  \centering
  \begin{tabular}{l|ccccccccc}
    \toprule
    \toprule
    \bf{Classification}
    & ImageNet & Caltech-101 & CIFAR-100 & SVHN & DTD & OxPet & OxFlowers & Eurosat & RESISC45 \\
    \midrule
    CLIP & 65.1 &	82.3 &	63.2 &	42.0 &	53.6 &	85.8 &	67.7 &	\bf{52.4} &	\bf{64.3} \\
    \STAIRs & 64.0 &	81.5 &	\bf{63.7} &	39.4 &	\bf{56.7} & \bf{85.9} &	67.0 & 49.5 &	62.2 \\
    \STAIRm & \bf{65.6} &	\bf{82.5} &	63.4 &	\bf{53.0} &	56.3 &	85.9 &	\bf{68.2} &	51.0 &	62.8 \\
    \bottomrule
  \end{tabular}
\end{subtable}
\label{tab:single_stage_retrieval}
\end{table}
%auto-ignore
\begin{table}[h]
  \caption{{\bf Interpretability of the STAIR models.} We report the top-K accuracy (\%) on ImageNet, CIFAR-100, and CalTech datasets, using all of the label among all of the tokens in the BERT vocabulary.}
  \vskip 0.15in
\small
      \centering
      \begin{tabular}{l|rr|rr|rr}
        \toprule
        & \multicolumn{2}{c|}{\bf ImageNet} & \multicolumn{2}{c|}{\bf CIFAR-100} & \multicolumn{2}{c}{\bf CalTech} \\
        & Top-1 & Top-100 & Top-1 & Top-100 & Top-1 & Top-100\\
        \midrule
        \STAIRm & 32.9 & 87.7 & 10.3 & 80.7 & 29.3 & 64.8\\
        \STAIRs & 0.0 & 0.6  & 0.0 & 0.3 & 0.0 & 0.2\\
        \bottomrule
      \end{tabular}
    
  \label{tab:single_stage_interp}
\end{table}

\section{Internal High-Quality Dataset}
\label{sec:ihqd}

The High Quality Image Text Pairs (a.k.a. HQITP-134m) dataset consists of approximately 134m diverse and high quality images paired with descriptive captions and titles. Images range in spatial resolution from 320 to 2048 pixels on the short side. All images are JPEG format and most are RGB. Each example image is associated with a title, and a list of several captions. A small fraction ($\ll 1\%$) of the examples are missing both captions and title. We favor the associated captions, and find that these tokenize to an average length of 20.1 tokens, although the shortest caption is only one token and the longest is over 1000. This dataset was licensed to our industrial research lab by a third party for commercial use.

\section{Our CLIP vs open-source CLIP}
In Section \ref{sec:training}, both the CLIP and STAIR models presented are trained with a batch size of 16,384. In order to demonstrate the correctness of our implementation, we train a CLIP-B/16 model and compare it with the open-source implementation~\footnote{ \url{https://huggingface.co/openai/clip-vit-base-patch16}}, denoted as CLIP\textsubscript{\textsc{open}}, using a doubled batch size of 32,768 in Table \ref{tab:openai_clip} We also attach the STAIR model metrics for reference. It shows our implemented CLIP performs competitively on ImageNet and significantly better on retrieval tasks. Furthermore, STAIR outperforms public CLIP on retrieval tasks despite half of the batch size.

\begin{table}[h]
\caption{{\bf Zero-shot transfer of open-source CLIP vs our CLIP using 32K batch size.} We report recall@K on Flickr30K and COCO, and top-1 accuracy (\%) on ImageNet. We attach our 16K batch-size STAIR metrics for reference.}
\vskip 0.15in
\small
    \centering
    \begin{tabular}{l|cccccc}
    \toprule
    & & \multicolumn{1}{c}{\bf ImageNet} & \multicolumn{2}{c}{\bf COCO 5K} & \multicolumn{2}{c}{\bf Flickr30K}\\
    & \multirow{2}* {Batch Size} & \multirow{2}*{Acc@1} & T2I & I2T & T2I & I2T \\
    & & & R@1 & R@1 & R@1 & R@1 \\
    \midrule
    CLIP\textsubscript{\textsc{open}} & 32,768 & \bf{68.6} & 33.3 & 54.1 & 63.5 & 82.3 \\
    CLIP   & 32,768 & 68.3 & 39.0 & 57.4 & \bf{67.0} & \bf{85.0}  \\   
    STAIR  & 16,384 & 65.6 & \bf{41.1} & \bf{57.7} & 66.6 & 81.2  \\ 
    \bottomrule
    \end{tabular}

    \label{tab:openai_clip}
    \vspace{-0.2cm}
\end{table}

\section{Token Based vs Embedding Based Search}
Developing embedding-based retrieval systems is emerging recently~\cite{hassantabar2021scann,faiss}. Despite the promising progress, deploying a large-scale embedding-based search system is still a big challenge because of several reasons. 
First of all, the embedding-based systems are often implemented using approximated neighbor search in the large-scale scenario, which usually involves k-means, and product quantization for clustering and partitioning. These operations are computationally costly in scale. Quantization is also often needed for saving disk costs but with dropped precision. Secondly, it usually requires extra computation or even re-computation for refreshing the index with changed data, as operations like k-means, and product quantization are data-dependent.

In contrast, a token-based retrieval system has no such problems because the indexing is through tokens themselves. In addition, the tokens in STAIR are optimized with FLOPs regularizer to encourage the uniform distribution in these sparse tokens, which is beneficial for retrieval. Besides, the interpretable token-based system gives extra benefits: 1) build customized query tree using logical operators; 2) leverage token-based blacklist/whitelist on both query and index side; 3) combine with other token-based futures in the inverted index system.

We also recognize the downside of current work --- STAIR is not able to capture high-level semantics. This can be potentially solved by combining dense embeddings and semantic tokens and we leave it as the future work.

\section{Image Prediction Weights}
\label{sec:image_token_examples_weights}
We list the detailed weights of predicted tokens for each image from Figure \ref{fig:image_token_examples} for reference.

\begin{table*}[ht!]
\centering
\small
    \begin{tabularx}{\textwidth}{ p{5.3cm}  |  p{5.3cm} |  p{5.3cm} }
        \hline
        
        \begin{minipage}{.28\textwidth}
        \includegraphics[trim=0 0 250 0,clip,width=\textwidth]{figures/wc3.pdf}
        \end{minipage}
        & 
        \begin{minipage}{.28\textwidth}
        \includegraphics[trim=0 0 255 0,clip,width=\textwidth]{figures/wc9.pdf}
        \end{minipage}
        & 
        \begin{minipage}{.28\textwidth}
        \includegraphics[trim=0 0 250 0,clip,width=\textwidth]{figures/wc1.pdf}
        \end{minipage}
        \\ 
        \#\#gul: 2.09, \#\#ll: 1.56, photograph: 1.35, plumage: 1.19, waves: 1.18, ashore: 1.13, beaches: 1.12, beach: 1.1, bird: 1.09, reintroduced: 1.08, version: 1.07, subspecies: 1.04, schleswig: 0.97, flick: 0.89, vague: 0.84, wave: 0.81, stillness: 0.79, turbulent: 0.79, republished: 0.78, coasts: 0.78
        &
        cake: 1.75, wedding: 1.47, cakes: 1.31, flick: 1.26, tier: 1.25, grapes: 1.23, fruit: 1.18, weddings: 1.15, receptions: 1.09, marriage: 1.0, \#\#lr: 0.99, \#\#illy: 0.98, slicing: 0.98, fruits: 0.88, berries: 0.87, marriages: 0.86, traditionally: 0.86, cheese: 0.86, lighted: 0.84, parties: 0.82
        &
        sydney: 1.98, australia: 1.92, australian: 1.42, opera: 1.33, kay: 1.32, \#\#dling: 1.27, paddle: 1.17, canoe: 1.16, harbour: 1.1, rowing: 1.04, competitors: 1.01, swimming: 0.97, regatta: 0.96, prosperous: 0.95, aquatics: 0.88, race: 0.85, inhabit: 0.85, waterfront: 0.79, pink: 0.79, wider: 0.78
        
        % extra blank line
        \\ \hline
        \begin{minipage}{.28\textwidth}
        \includegraphics[trim=0 0 252 0,clip,width=\textwidth]{figures/wc4.pdf}
        \end{minipage}
        & 
        \begin{minipage}{.28\textwidth}
        \includegraphics[trim=0 0 250 0,clip,width=\textwidth]{figures/wc5.pdf}
        \end{minipage}
        & 
        \begin{minipage}{.28\textwidth}
        \includegraphics[trim=0 0 252 0,clip,width=\textwidth]{figures/wc11.pdf}
        \end{minipage} \\
        airport: 1.6, airline: 1.35, airlines: 1.22, window: 1.22, airports: 1.21, delta: 1.21, nexus: 1.06, lax: 0.99, concourse: 0.97, \#\#fk: 0.92, gate: 0.87, flights: 0.86, traveled: 0.84, view: 0.82, airways: 0.79, aircraft: 0.76, mildly: 0.76, \#\#lr: 0.75, rainy: 0.75, traveling: 0.75
        &
        sheep: 1.74, herd: 1.54, goats: 1.36, cumbria: 1.32, flick: 1.29, livestock: 1.29, driving: 1.27, goat: 1.24, scotland: 1.09, traffic: 1.04, lamb: 1.02, loch: 1.01, tyrol: 0.94, skye: 0.92, unusual: 0.92, flock: 0.91, animals: 0.9, heading: 0.89, roadside: 0.87, inaccessible: 0.85
        &
        nippon: 2.26, ana: 2.03, airline: 1.7, aircraft: 1.4, airlines: 1.37, 77: 1.36, airport: 1.3, el: 1.16, flights: 1.09, departing: 1.06, \#\#7: 1.01, republished: 0.92, busiest: 0.91, boeing: 0.9, livery: 0.88, aviation: 0.87, creditors: 0.85, spotting: 0.8, japan: 0.75, commons: 0.72,

        % extra blank line
        \\
        \hline
        \begin{minipage}{.28\textwidth}
        \includegraphics[trim=0 0 252 0,clip,width=\textwidth]{figures/wc7.pdf}
        \end{minipage}
        & 
        \begin{minipage}{.28\textwidth}
        \includegraphics[trim=0 0 250 0,clip,width=\textwidth]{figures/wc8.pdf}
        \end{minipage}
        & 
        \begin{minipage}{.28\textwidth}
        \includegraphics[trim=0 0 252 0,clip,width=\textwidth]{figures/wc10.pdf}
        \end{minipage} \\
        abbey: 2.7, westminster: 2.43, nw: 2.03, \#\#minster: 1.87, northwest: 1.82, west: 1.78, ab: 1.67, \#\#bey: 1.62, roads: 1.6, \#\#3: 1.59, \#\#ey: 1.56, road: 1.51, abby: 1.38, sign: 1.23, minister: 1.22, peterborough: 1.21, \#\#mist: 1.13, nwa: 1.0, agreed: 0.97, n: 0.94
        &
        geese: 1.86, goose: 1.75, canada: 1.4, photograph: 1.31, mating: 1.18, ducks: 1.18, \#\#oides: 1.14, brant: 1.13, subspecies: 1.12, wildlife: 1.11, awhile: 1.1, courtship: 1.08, version: 1.04, swans: 0.97, lake: 0.97, pond: 0.95, sired: 0.95, synchronized: 0.94, duck: 0.94, flick: 0.93
        &
        thanksgiving: 1.72, turkey: 1.46, roast: 1.26, very: 1.25, pie: 1.23, christmas: 1.04, \#\#cans: 1.02, mmm: 1.01, chicken: 0.99, holidays: 0.88, feast: 0.82, republished: 0.81, stuffing: 0.78, dinner: 0.76, bread: 0.76, tasted: 0.75, convinces: 0.75, televised: 0.74, meal: 0.73
        \\ \hline
    \end{tabularx}
    \caption{Qualitative analysis of the sparse embedding}
    \label{tbl:myLboro}
\end{table*}

%%%%%%%%%%%%%%%%%%%%%%%%%%%%%%%%%%%%%%%%%%%%%%%%%%%%%%%%%%%%%%%%%%%%%%%%%%%%%%%
%%%%%%%%%%%%%%%%%%%%%%%%%%%%%%%%%%%%%%%%%%%%%%%%%%%%%%%%%%%%%%%%%%%%%%%%%%%%%%%

\end{document}